%% file: main.tex
\documentclass[10pt,twocolumn]{article}

\usepackage[letterpaper,margin=0.82in,columnsep=0.26in]{geometry}

\usepackage[T1]{fontenc}
\usepackage[utf8]{inputenc}
\usepackage{amsmath}
\usepackage{amssymb}
\usepackage{amsthm}
\usepackage{mathtools}
\usepackage{newtxtext,newtxmath}
\usepackage{microtype}

\usepackage{booktabs}
\usepackage{array}
\usepackage{multirow}
\usepackage{graphicx}
\usepackage{float}
\usepackage{xcolor}
\usepackage{caption}
\usepackage{subcaption}
\usepackage{enumitem}
\usepackage{balance}
\usepackage{algorithm}
\usepackage{algpseudocode}

\usepackage{tikz}
\usetikzlibrary{arrows.meta,positioning,shapes.geometric,fit,backgrounds,calc}
\usepackage{pgfplots}
\pgfplotsset{compat=1.18}
\usepgfplotslibrary{groupplots}

\usepackage[colorlinks=true,linkcolor=auburl,citecolor=auburl,
            urlcolor=auburl,breaklinks=true]{hyperref}
\usepackage{cleveref}

\definecolor{auburl}{HTML}{0E7C86}
\definecolor{aubnavy}{HTML}{123640}
\definecolor{aubteal}{HTML}{CFEAE9}
\definecolor{auborange}{HTML}{E2663B}
\definecolor{aubpeach}{HTML}{FBEBE3}
\definecolor{aubgreen}{HTML}{2E9E6B}
\definecolor{aubsand}{HTML}{FBF1DD}
\definecolor{aubgrey}{HTML}{5B6670}
\definecolor{aubline}{HTML}{DEE3E6}

\theoremstyle{definition}
\newtheorem{definition}{Definition}
\newtheorem{assumption}{Assumption}
\theoremstyle{plain}
\newtheorem{proposition}{Proposition}
\newtheorem{corollary}{Corollary}

\crefname{definition}{Definition}{Definitions}
\Crefname{definition}{Definition}{Definitions}
\crefname{assumption}{Assumption}{Assumptions}
\Crefname{assumption}{Assumption}{Assumptions}
\crefname{proposition}{Proposition}{Propositions}
\Crefname{proposition}{Proposition}{Propositions}
\crefname{corollary}{Corollary}{Corollaries}
\Crefname{corollary}{Corollary}{Corollaries}

\newcommand{\sys}{Augur}
\newcommand{\code}[1]{\texttt{\small #1}}
\newcommand{\actset}{\mathcal{A}}
\DeclareMathOperator*{\argmax}{arg\,max}
\DeclareMathOperator{\clip}{clip}
\DeclareMathOperator{\sign}{sign}
\DeclareMathOperator{\rel}{rel}
\DeclareMathOperator{\Var}{Var}

\DeclareMathOperator{\anc}{anc}
\newcommand{\indic}[1]{\mathbf{1}\!\left[#1\right]}
\newcommand{\bigO}{\mathcal{O}}
\newcommand{\pp}{\,\text{pp}}

\input{figures.tex}

\begin{document}

\title{\sys{}: A Synthetic Decision Lab for Rehearsing Reactions to
Product and Policy Changes}
\author{Rahul Khedar \quad Mayank Malhotra \quad Avinash Karn\\[5pt]
\normalsize PayPal AI}
\date{}
\maketitle

\begin{abstract}
Before a product or policy change ships, the question that matters is how people
will react to it. \sys{} rehearses that reaction offline: it builds a typed
knowledge graph from the change documents, populates a grounded persona market,
simulates the interaction, and returns an auditable decision memo recommending
one of five actions. We assemble \emph{Gold-50}, fifty real product and policy
episodes whose real-world outcome is known, adjudicated against the public
record, and score the five-way release verdict against it.

Our central finding is methodological and negative: most of the measured gap
between frontier cloud models and open-weight models we fine-tune and serve
offline is attributable to an under-specified evaluation, not a difference in
capability. We show this three ways. First, the prompt
envelope alone can dominate the score: holding weights, cases and scorer fixed,
one system---a LoRA-SFT adapter on Qwen3-32B---swings from $0\%$ to $73\%$. Second, in a matched $2\times2$
ablation, defining the decision taxonomy in the prompt---with no model
change---lifts every frontier model by $+24$ to $+34\pp$; under the
under-specified prompt, Qwen3-32B LoRA-SFT served offline \emph{beats} all three frontier
models (paired McNemar, Holm-corrected), and once the prompt is fair no
significant difference from any of them is detected.
Third, agreement with the distillation teacher rises without accuracy following,
and the full pipeline amplifies a systematic ``over-doom'' bias
rather than improving the verdict. Separately, we validate the reaction layer on its own
terms: blind judges across four model families find the synthetic reaction
recovers $67$--$90\%$ of the concerns the public actually raised, and a
pre-registered ablation locates its value---largest where the decision is
hardest, redundant near ceiling. The pipeline that
regenerates every number and figure here is available from the authors.
\end{abstract}

\section{Introduction}
\label{sec:intro}

Organizations decide whether to ship changes---a pricing update, a policy
revision, a feature launch---under uncertainty about how the people affected will
react. The evidence they would most like to have, the reaction itself, does not
exist until after the decision is made. \sys{} is a system for manufacturing a
usable proxy for that evidence \emph{before} the fact: it reads the documents
that define a change, constructs a market of stakeholders grounded in those
documents, simulates their interaction~\cite{park2023generative,li2023camel}, and returns a decision memo that
recommends one of five release actions and shows its work.

A system like this is only as trustworthy as the measurement used to judge it,
and this is where the paper concentrates its effort. It is easy to run a fleet of
models through a decision harness, read off a leaderboard, and conclude that
frontier cloud models are far ahead of anything one can fine-tune and serve
offline. We built that leaderboard, and then spent most of our effort showing
that the conclusion it invites is largely an artifact of \emph{how} the models
were asked, not of what they can do.

\paragraph{Aim.} Our goal is to establish, on a benchmark of real decisions with
verified outcomes, how much of the frontier-versus-offline gap on this task is
real capability and how much is measurement, and to give a mechanism for the
difference. The answer, \emph{on these fifty cases}, is that the gap is mostly
measurement: once the decision taxonomy is defined in the prompt for everyone, a
Qwen3-32B open-weight model served offline shows no statistically significant
difference from the frontier models tested.

\paragraph{Contributions.}
\begin{enumerate}[leftmargin=1.2em,itemsep=1pt,topsep=1pt]
\item \textbf{Gold-50}, a benchmark of $50$ real product and policy episodes with
manually verified real-world outcomes, balanced ten per class across a five-way
action taxonomy, together with the \sys{} system that produces auditable verdicts
on it (\Cref{sec:system,sec:goldset}).
\item \textbf{The prompt envelope can dominate the result.} With weights, cases
and scoring code held fixed, changing only the prompt moves one system---the
deployed Qwen3-32B LoRA-SFT (\code{sft\_v3})---from
$0\%$ to $73\%$ (\Cref{sec:harness}). Any single-prompt comparison of this task
is therefore uninterpretable, and we report the compliance rate alongside every
accuracy.
\item \textbf{A matched $2\times2$ ablation resolves most of the gap}
(\Cref{sec:debias}). On the same fifty report-stage contexts through the same
direct harness, appending a shared taxonomy-definition block lifts every frontier
model by $+24$ to $+34\pp$. Under the under-specified prompt the offline Qwen3-32B LoRA-SFT
significantly beats all three frontier models (Holm-corrected); under the fair
prompt no statistically significant difference from any of them is detected. The
mechanism differs by family, so the two
interventions are complementary rather than redundant.
\item \textbf{Two failure modes with a formal account.} Agreement with the
distillation teacher climbs from $10\%$ to $70\%$ across fine-tuning rounds with
no matching movement in outcome accuracy (\Cref{sec:teacher}), and the full
five-stage pipeline amplifies an over-doom bias, scoring at or below the
five-way chance rate (\Cref{sec:pipeflow}). \Cref{sec:appendix-props} proves why
both are expected: distillation collapses judge independence, and any
exact-match scorer imposes a schema-compliance ceiling.
\item \textbf{The reaction layer is faithful, and its value is located, not flat.}
Blind judges across four model families find the synthetic reaction anticipates
$67$--$90\%$ of the concerns the public actually raised (\Cref{sec:fidelity}), and
a \emph{pre-registered} matched with/without-simulation ablation
(\Cref{sec:simablation}) shows the report-stage reaction signal helps most exactly
where the title-only baseline is weakest ($+20\pp$ for the weakest solver) and is
redundant near ceiling---so the over-doom of the previous point is a property of
\emph{stacking} the generative stages, not of the reaction evidence itself.
\item \textbf{A reproducible artifact-to-PDF pipeline.} Every number, table cell
and plotted mark in this paper is generated from the run artifacts and then
re-asserted against them; nothing is transcribed by hand (\Cref{sec:repro}).
\end{enumerate}

\section{The Augur System}
\label{sec:system}

\begin{figure*}[t]
\centering
\begin{tikzpicture}[
  font=\small,
  box/.style={rectangle,rounded corners=4pt,draw=auburl,fill=aubteal,
    text=aubnavy,minimum height=1.55cm,text width=2.35cm,align=center,
    inner sep=3pt,line width=0.8pt},
  doc/.style={rectangle,rounded corners=4pt,fill=aubnavy,text=white,
    minimum height=1.55cm,text width=1.8cm,align=center,inner sep=3pt},
  inter/.style={rectangle,rounded corners=4pt,draw=auborange,fill=aubpeach,
    text=aubnavy,minimum height=1.2cm,text width=6.2cm,align=center,
    dash pattern=on 4pt off 2pt,line width=0.8pt},
  flow/.style={-{Stealth[length=3mm]},line width=1.1pt,auburl},
  back/.style={-{Stealth[length=3mm]},line width=1pt,auborange,
    dash pattern=on 4pt off 2pt},
  num/.style={circle,fill=auburl,text=white,inner sep=1.3pt,
    font=\scriptsize\bfseries}
]
\node[doc] (d) {\textbf{Documents}\\[2pt]\scriptsize PDF \textperiodcentered MD \textperiodcentered text};
\node[box,right=0.7cm of d] (g) {\textbf{Graph Build}\\[2pt]\scriptsize $G(D)$:
  ontology + entities};
\node[box,right=0.7cm of g] (e) {\textbf{Environment}\\[2pt]\scriptsize $P(G,q)$:
  personas + stance};
\node[box,right=0.7cm of e] (s) {\textbf{Simulation}\\[2pt]\scriptsize $S$:
  OASIS rounds};
\node[box,right=0.7cm of s] (r) {\textbf{Report}\\[2pt]\scriptsize $R$: evidence
  + forecast $F$};
\node[right=0.7cm of r,text=aubgreen,font=\bfseries,align=center] (out)
  {$a^\star$\\Decision};
\node[num] at (g.north west) {1}; \node[num] at (e.north west) {2};
\node[num] at (s.north west) {3}; \node[num] at (r.north west) {4};
\draw[flow] (d) -- (g);  \draw[flow] (g) -- (e);  \draw[flow] (e) -- (s);
\draw[flow] (s) -- (r);  \draw[flow] (r) -- (out);
\node[inter,below=1.35cm of s] (i)
  {\textbf{\tikz[baseline=(n5.base)]\node[circle,fill=auborange,text=white,
     inner sep=1.3pt,font=\scriptsize\bfseries](n5){5};~~Interaction}\\[2pt]\scriptsize
  interview any persona \textperiodcentered ask ``why did you react?'' \textperiodcentered re-run};
\draw[back] (i.west) -| (g.south);
\draw[back] (i.east) -| (r.south);
\end{tikzpicture}
\caption{The \sys{} pipeline; symbols are the operators of \Cref{eq:pipeline}.
Each forward stage persists an inspectable artifact; the interaction stage feeds
findings back into the graph or report.}
\label{fig:pipeline}
\end{figure*}
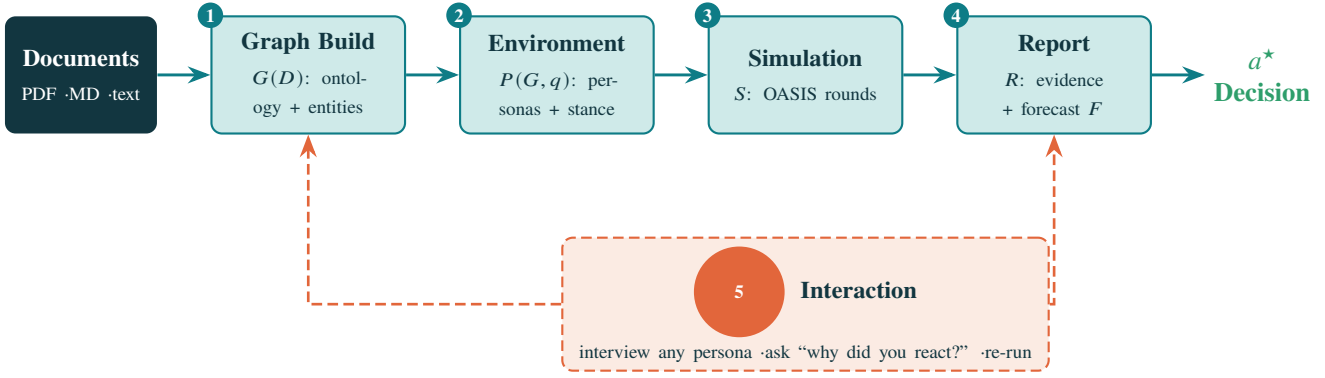

\sys{} factors a decision into five single-responsibility stages, each of which
reads the artifacts of the stages before it and writes its own. The forward path
is $\text{documents}\to\text{graph}\to\text{personas}\to\text{simulation}\to
\text{report}$, and a fifth interaction stage supports post-hoc interrogation of
any persona. \Cref{fig:pipeline} shows the flow; \Cref{sec:appendix-formal}
gives the formal operators, and we summarize them here.

\paragraph{Stages.} \emph{Graph build} infers a typed ontology from the change
documents and extracts a directed knowledge graph of entities and relations.
\emph{Environment} synthesizes a persona population grounded in that graph---each
persona references graph entities, so the market is specific to the change rather
than a generic crowd. \emph{Simulation} runs the personas for several rounds on an OASIS-style engine~\cite{yang2024oasis},
producing a timeline of posts, replies and reactions with peer-influenced
sentiment. \emph{Report} retrieves graph-expanded evidence~\cite{lewis2020rag,edge2024graphrag}, writes a forecast,
and emits the recommended action $a^\star$ from the five-way taxonomy
\[
\actset=\{\textsc{ship},\textsc{revise},\textsc{delay},\textsc{segment},\textsc{mitigate}\}.
\]
Between stages, three pure-code auditors read the finished transcript and can
each buy at most one corrective round under a global budget, and a cross-family
critic gates the verdict.

\begin{figure}[tbp]
\centering
\begin{tikzpicture}[
  font=\footnotesize,
  comp/.style={rectangle,rounded corners=3pt,draw=auburl,fill=white,
    text=aubnavy,align=center,minimum height=0.85cm,inner sep=3pt,
    line width=0.7pt,font=\scriptsize},
  scomp/.style={comp,draw=aubgreen},
  stor/.style={comp,draw=auborange,fill=aubsand},
  llm/.style={comp,draw=aubnavy,fill=aubteal!30},
  down/.style={-{Stealth[length=2.6mm]},line width=1pt,aubnavy},
  lab/.style={font=\scriptsize\bfseries,text=aubgrey}
]
\node[comp,minimum width=3.1cm] (front) {Flask + Jinja frontend};
\node[lab,above left=-2pt and -2.6cm of front.north] {PRESENTATION};
\node[comp,below left=0.9cm and 0cm of front.south,anchor=north west,xshift=-1.1cm] (r1) {Graph\\routes};
\node[comp,right=0.25cm of r1] (r2) {Sim.\\routes};
\node[comp,right=0.25cm of r2] (r3) {Report\\routes};
\node[lab] at ($(r1.north west)+(0.9,0.28)$) {API \textperiodcentered ORCHESTRATOR};
\node[scomp,below=1.15cm of r1.south,anchor=north,xshift=-0.15cm] (s1) {NER +\\Ontology};
\node[scomp,right=0.2cm of s1] (s2) {OASIS\\sim};
\node[scomp,right=0.2cm of s2] (s3) {Graph\\memory};
\node[scomp,right=0.2cm of s3] (s4) {Graph\\tools};
\node[lab] at ($(s1.north west)+(1.0,0.28)$) {SERVICES \textperiodcentered AGENTS};
\node[stor,below=1.15cm of s1.south,anchor=north] (st1) {GraphStorage\\\scriptsize interface};
\node[stor,right=0.3cm of st1] (st2) {Neo4j 5.18\\\scriptsize swappable};
\node[lab] at ($(st1.north west)+(1.0,0.28)$) {STORAGE};
\node[llm,right=0.55cm of st2] (l1) {Gateway /\\Hosted};
\node[llm,right=0.2cm of l1] (l2) {Local\\runtime};
\node[lab] at ($(l1.north west)+(0.9,0.28)$) {LLM BOUNDARY};
\draw[down] (front.south) -- ++(0,-0.32) -| (r2.north);
\draw[down] (r2.south) -- ++(0,-0.30) -| (s2.north);
\draw[down] (s1.south) -- ++(0,-0.30) |- (st1.north);
\draw[{Stealth[length=2.6mm]}-{Stealth[length=2.6mm]},line width=1pt,auborange]
   (st1.east) -- (st2.west);
\draw[down] (s4.south) -- ++(0,-0.30) -| (l1.north);
\end{tikzpicture}
\caption{Layered architecture. A Flask/Jinja frontend drives an API and
orchestrator layer; the single-responsibility agents (named-entity recognition [NER] and ontology, OASIS
simulation, graph memory, graph tools) sit below it and interact with the rest of
the system only through an abstracted \code{GraphStorage} interface (Neo4j today,
swappable) and one OpenAI-compatible model boundary (a hosted gateway or a local
runtime). Nothing above the storage or model boundary knows which backend is in
use, which is what makes the fully offline deployment of \Cref{sec:offline} a
configuration change.}
\label{fig:arch}
\end{figure}
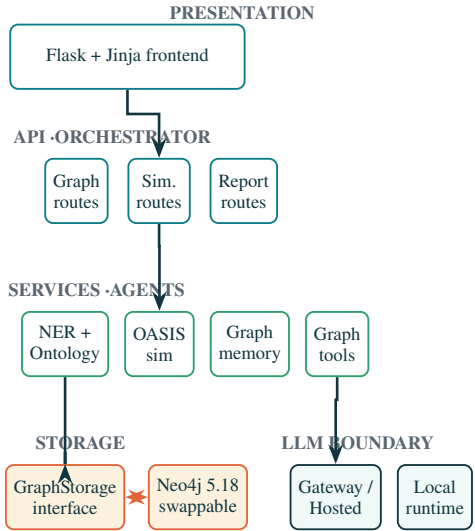

\paragraph{Engineered independence.}
\label{sec:autonomy}
Agents never call one another; each reads persisted artifacts and writes its own,
and a derived-from dependency directed acyclic graph (DAG) makes regeneration of any artifact cascade to
everything downstream (\Cref{sec:appendix-formal}). The ensemble that produces
the verdict is drawn \emph{cross-family}: the critic and cross-check come from a
model family other than the report author's~\cite{panickssery2024llm}. \Cref{sec:appendix-props} explains
why this matters---same-family voting forfeits most of the amplification an
ensemble can provide, and it is exactly the property that distillation destroys.
\Cref{fig:arch} shows the layered structure that enforces this: every agent
reaches the rest of the system only through an abstracted storage interface and a
single OpenAI-compatible model boundary, so the graph store (Neo4j today) and the
model backend (a hosted gateway or a local runtime) are both swappable without
touching agent code---the property that makes the fully offline deployment of
\Cref{sec:offline} a configuration change rather than a rewrite.

\paragraph{Offline deployment.}
\label{sec:offline}
The system is designed to run without a cloud dependency. We fine-tune the
generation stages with low-rank adaptation (LoRA)~\cite{hu2022lora} on distilled traces~\cite{hinton2015distilling,kim2016sequence} and serve them on a single
accelerator; the untuned cross-family critic is kept off-the-shelf so that its
errors are not the generator's errors. Training details, adapters and serving
cost are in \Cref{sec:appendix-ft}.

\section{The Gold-50 Benchmark}
\label{sec:goldset}

Judging a decision system requires cases whose \emph{realized outcome is known}.
Gold-50 is fifty real product and policy episodes---each a change that a real
organization actually made---for which the real-world outcome was later observed
and can be mapped onto the five-way action taxonomy. Each label records the
\emph{realized disposition} of the episode (what actually happened), not an
abstract ``optimal action.'' Every one of the $50$ labels
was adjudicated in a single author pass against the public record, assisted by
Claude Opus~4.8 (adjudication run 2026-09-06), with a median of three independent
sources per case. Because Claude Opus~4.8 also appears as one of the benchmarked
frontier models, its adjudication role is a potential source of circularity in
the frontier-model comparison; we disclose it here and read the Opus~4.8 Gold-50
numbers with that caveat. The labeling convention for each of the five classes is
given in \Cref{tab:convention}, and the full per-case ledger---label, outcome
driver, year and source count---is in \Cref{tab:goldprov}. Each case is annotated
with the mechanism that actually decided its outcome, so that accuracy can be
stratified by \emph{why} a decision went the way it did. Because the labels are a
single-pass author adjudication rather than a multi-annotator consensus, we treat
Gold-50 as an author-adjudicated retrospective benchmark and scope every claim to it.

Every system is shown only the \emph{pre-decision} information for a case---the
policy or product document as it stood before the outcome was known. The realized
outcome, the outcome-driver annotation and the source list are held out of the
model input entirely and are used only for scoring, so no system can read the
answer off its own prompt.

The set is balanced at ten cases per class across
$\{\textsc{ship},\textsc{revise},\textsc{delay},\textsc{segment},\textsc{mitigate}\}$,
which fixes the five-way chance rate at exactly $20\%$ and makes a raw accuracy
directly comparable to it. The verification annotations sort the cases into four
outcome drivers: public \emph{reaction} ($28$ cases), \emph{regulatory} action
($11$), engineering \emph{readiness} ($7$), and \emph{business} performance
($4$). The reaction stratum is both the largest and the one the pipeline is
architecturally aimed at, and---as \Cref{sec:appendix-extra} shows---the one every
system does worst on, which is the honest limit of a reaction-simulating design.

Because the target action is parsed out of a structured field, we score two ways
throughout: a conservative accuracy that counts any unparseable output as wrong,
and the accuracy conditional on in-schema output, reported with its parsed count.
\Cref{sec:appendix-props} makes precise why both are necessary.

\section{Experimental Setup}
\label{sec:setup}

\paragraph{Systems.} We evaluate eleven systems on the report stage: three
frontier cloud models (Claude Sonnet~5, Claude Opus~4.8, GPT-5.2) and eight
open-weight configurations spanning $1.7$B to $32$B parameters, both untuned
bases and their adapters from LoRA supervised fine-tuning (SFT), plus one variant extended with direct preference optimization (DPO), the SFT\ensuremath{\to}DPO configuration~\cite{rafailov2023dpo}. The base
backbones are Qwen3-1.7B, Qwen3-8B and Qwen3-32B~\cite{qwen3}, Llama-3-8B-Instruct~\cite{llama3} and
Gemma-3-27B-it~\cite{gemma3}; each named result below identifies its exact configuration, and
\Cref{tab:gold50} lists all eleven. The deployed open checkpoint is a LoRA-SFT
adapter on Qwen3-32B (\code{sft\_v3}), and its preference-tuned sibling is
Qwen3-32B SFT\ensuremath{\to}DPO (\code{dpo}); their lineage and hyperparameters
are in \Cref{sec:appendix-ft}.

\paragraph{Harness and prompt arms.} Every comparison in this paper holds the
fifty cases and the scoring code fixed and varies only the model and the prompt.
The \emph{deployed} report-stage prompt is the one the application serves. The
\emph{debias} arm appends a single shared block that defines each of the five
actions and discourages hedging. The \emph{abstract} arm, used only as a
schema-compliance probe (\Cref{sec:harness}), asks for a verdict through a
generic instruction prompt. No arm changes model weights.

\paragraph{Statistics.} Wilson score intervals give the $95\%$ confidence band on
each accuracy~\cite{wilson1927}; at $n{=}50$ these are wide and overlap for most
pairs, so we do not compare systems by eyeballing intervals. Because every
system sees the \emph{same} cases, the correct paired test is the two-sided exact
McNemar test on the discordant pairs~\cite{mcnemar1947}, which we use for every
head-to-head claim. Where a claim rests on a family of paired tests---the
leader-versus-frontier comparisons under each prompt arm, and the fine-tuning
deltas---we report Holm-corrected and Benjamini-Hochberg-corrected
$p$-values~\cite{holm1979,benjamini1995} alongside the raw ones; our primary
hypothesis family is the leader-versus-frontier comparison under the shared-taxonomy
(debias) prompt. A non-significant McNemar result at $n{=}50$ is a failure to
detect a difference, not evidence of equivalence: the sample is too small to
support an equivalence test, and we never read ``no significant difference'' as
``the same.'' Class-level performance is summarized by macro-F1.

\paragraph{Data hygiene.} Gold-50 is an evaluation-only set and is disjoint from
everything used to build the open checkpoints. The fine-tuning corpus is
synthetic persona-generation traces; it shares no case identifier with Gold-50
($0$ overlapping ids), no case document matches any training document, and the
strongest cross-set link we find is a single shared company name in at most one
case---consistent with two independently constructed corpora that happen to
mention the same well-known organization. The audit is recomputed from the raw
splits by the metrics pipeline and reported in \Cref{sec:appendix-repro};
no Gold-50 case was used for SFT, DPO, validation or model selection.

\section{Results}
\label{sec:results}

\subsection{The report-stage leaderboard}
\label{sec:leaderboard}

\Cref{tab:gold50,fig:leaderboard} give the report-stage leaderboard under the
deployed prompt. The deployed Qwen3-32B LoRA-SFT (\code{sft\_v3}) leads at $31/50=62.0\%$ (Wilson
$[48.2,74.1]$, macro-F1 $0.59$); the Qwen3-32B SFT\ensuremath{\to}DPO variant matches it, and the
untuned Qwen3-8B base reaches $54.0\%$. At the bottom, the Qwen3-1.7B LoRA-SFT adapter collapses to
$6.0\%$ with $35/50$ unparseable outputs---a compliance failure, not a judgment
failure, and a first concrete instance of the ceiling of \Cref{sec:appendix-props}.

\tabGoldLeaderboard
\figLeaderboard

The Wilson intervals overlap for nearly every pair, so we decide separation with
the paired test. Exactly three systems are separated from the leader (the
Qwen3-32B LoRA-SFT) at $\alpha{=}0.05$: the Qwen3-1.7B LoRA-SFT, the
Llama-3-8B-Instruct base, and the Gemma-3-27B-it base; the second-place
Qwen3-32B SFT\ensuremath{\to}DPO variant shows zero discordant pairs with the leader,
and the $60\%$ Llama-3-8B LoRA-SFT adapter shows no significant difference from it. In other words, the
top of the leaderboard is a cluster, not a ranking, at this sample size.

\paragraph{Fine-tuning is not uniformly beneficial.} Applying one recipe and one
dataset to five backbones (\Cref{tab:paired,fig:paired}) produces effects that
range from decisive to destructive: $+44\pp$ on Llama-3-8B ($p=3.0\times10^{-6}$)
and $-40\pp$ on Qwen3-1.7B ($p=1.1\times10^{-5}$), with the three mid-size
effects not significantly different from noise at $n{=}50$. The gain on Llama-3-8B comes from
the SFT fixing $23$ cases while breaking $1$; the collapse on the smallest backbone,
Qwen3-1.7B, is the mirror image, breaking $21$ while fixing $1$, as the adapter degrades the
model's already-fragile schema compliance.

\tabPaired
\figPaired

\subsection{The prompt envelope can dominate the result}
\label{sec:harness}

Before comparing systems, we have to establish how much of a measured difference
is judgment and how much is the harness. \Cref{fig:formatood} answers this on an
answer-keyed subset: the same weights and the same cases, scored the same way,
under two prompts. Through the deployed report-stage prompt the three checkpoints
(Qwen3-32B base, Qwen3-8B LoRA-SFT, and the deployed Qwen3-32B LoRA-SFT) score
$54.5$--$72.7\%$; through a generic abstract prompt the same weights emit actions
outside the five-way schema and accuracy collapses. The fine-tuned Qwen3-32B
LoRA-SFT is the \emph{most} harness-sensitive of the three, because SFT bound its output format
tightly to the serving prompt: it is the worst under the abstract prompt and the
best under the deployed one, swinging from $0\%$ to $73\%$.

\figFormatOOD

This is not a curiosity---it is the reason a single-prompt leaderboard on this
task cannot be trusted~\cite{sclar2024quantifying,mizrahi2024state}, and it motivates the matched ablation of
\Cref{sec:debias}. We therefore report all headline numbers under the deployed
prompt and treat the abstract prompt as a schema-compliance probe rather than a
judgment measurement.

\subsection{The debias 2$\times$2: most of the gap is the prompt}
\label{sec:debias}

The cleanest test of the paper's thesis puts frontier and open-weight models on
identical footing: the \emph{same} fifty report-stage contexts, the \emph{same}
direct harness, and a single binary intervention---whether the shared
taxonomy-definition block is appended. \Cref{fig:debias2x2} and
\Cref{tab:debias2x2} give the full grid; \Cref{tab:debiash2h} settles the
head-to-head with the paired test.

\figDebiasGrid

Defining the taxonomy, with \emph{no model change}, lifts every frontier model by
$+24$ to $+34\pp$, and every frontier gain is significant at $p<0.01$: Sonnet~5
$46\to80\%$, Opus~4.8 $44\to68\%$, GPT-5.2 $38\to66\%$. The offline Qwen3-32B LoRA-SFT
moves $62\to78\%$ and its DPO variant (Qwen3-32B SFT\ensuremath{\to}DPO) $62\to76\%$. The low frontier scores under
the deployed prompt were therefore largely a prompt artifact, not a capability
gap.

The head-to-head makes the consequence precise. Under the under-specified
baseline prompt the offline Qwen3-32B LoRA-SFT significantly \emph{beats} all three
frontier models ($p=0.039/0.012/0.002$; all significant after Holm correction,
\Cref{tab:debiash2h}), winning $10$--$13$ discordant cases
against $1$--$2$. Once the taxonomy is defined for everyone, no statistically
significant difference from any of the three is detected
($p=1.000/0.227/0.180$; Holm-corrected in \Cref{tab:debiash2h}) --- a failure to
detect a difference at $n{=}50$, not evidence of equivalence. The Qwen3-32B LoRA-SFT
served offline thus shows no significant difference from the frontier when the
prompt is fair and wins whenever it is not.

\tabDebiasHead

\paragraph{The mechanism differs by family.} The two interventions---fine-tuning
and the prompt block---are not redundant, because they fix different failures
(\Cref{fig:debiasmech}). For the frontier models the block unwinds a genuine
\emph{mitigate}-collapse: over-prediction of \emph{mitigate} against a true
frequency of $10/50$ falls sharply (Sonnet $36\to16$, Opus $33\to18$, GPT
$35\to19$). The Qwen3-32B fine-tunes (\code{sft\_v3}, \code{dpo}) never had that collapse (\code{sft\_v3} $14\to16$);
their gain comes instead from recovering the \emph{segment} class, which was
nearly unreachable at $2/10$ recall and rises to $8/10$ once the taxonomy names
what \emph{segment} means. Two distinct failures, two complementary fixes.

\figDebiasMech

\subsection{The full pipeline amplifies over-doom}
\label{sec:pipeflow}

Adding the simulation stages ahead of the report does not improve the verdict;
it makes it worse. \Cref{fig:pipebias,tab:cloudflow} run the \emph{complete}
five-stage pipeline over the same fifty documents in three configurations: every
stage on Claude Opus~4.8, every stage on GPT-5.2, and the system's default
\emph{per-agent} routing (\code{auto}), which assigns each stage the model best
suited to it---Sonnet~5 for the high-volume extraction and simulation stages,
Opus~4.8 for persona and report synthesis, and a cross-family GPT-5.2 critic so
the audit stays independent. Opus
scores $40.0\%$ and GPT $36.0\%$---below every report-only system above eighth
place---and the default per-agent \code{auto} configuration scores $14.0\%$,
\emph{below} the $20\%$ five-way chance rate. The mechanism is a systematic over-doom bias: the
full pipeline answers \emph{mitigate} on $68\%$, $80\%$ and $62\%$ of cases
against a true frequency of $20\%$, and the GPT run never once recommends
\textsc{ship}.

\figPipeBias
\tabCloudFlow

This comparison is confounded by model family (the full-pipeline runs are
frontier, the report-only baselines open-weight), so we read it as evidence about
the \emph{bias} the simulation stages induce rather than as a clean accuracy
ranking; \Cref{sec:debias} already isolated the prompt component of the gap. The
practical reading is that more machinery is not more accuracy on this task, and
the extra stages cost latency---a median of several minutes per case
(\Cref{tab:cloudflow}). Whether the harm is the reaction \emph{signal} or the
\emph{stacking} of stages that carry it is exactly what the matched ablation of
\Cref{sec:simablation} decides.

\subsection{The reaction simulation is a faithful proxy}
\label{sec:fidelity}

\Cref{sec:pipeflow} asks whether adding the simulation improves the \emph{verdict};
for a tool whose purpose is to rehearse reaction \emph{before} a decision ships, the
prior question is whether the synthetic reaction it produces resembles the real one at
all. At decision time no real reaction exists, so the synthetic transcript is the
production artifact and its faithfulness is the construct the whole system rests on. We
test it directly on the $16$ cases---a subset of the $28$-case reaction stratum
(\Cref{fig:drivers})---for which a real, outcome-scrubbed contemporaneous
public reaction could be sourced. Each reaction block---the
exact personas-and-timeline text the report stage consumes---is rated \emph{blind}: the
judge sees one block at a time, is never told whether it is synthetic or real, and never
sees the outcome. Four judges spanning families do the rating (cloud: GPT-5.2,
Opus~4.8; open-weight: Qwen3-32B, Gemma-3-27B-it), which also yields an inter-judge
reliability check on the measurement itself. The load-bearing metric is
\emph{concern recall}: the fraction of the concerns raised in the real reaction that the
synthetic reaction anticipated.

\tabFidelity

Across all four judges the synthetic reaction anticipates $67$--$90\%$ of the concerns
the public actually raised (\Cref{tab:fidelity}), and the two open-weight judges score it
\emph{highest} ($86$--$90\%$), so the result is not a cloud-judge artifact. Inter-judge
reliability is high---across the six judge pairs, stance ratings agree within one point on
$75$--$100\%$ of cases (the two cloud judges, Opus and GPT, agree on all $16$)---so the
metric is stable, not a single model's idiosyncrasy. We are deliberately conservative
about the weaker signals. Direction agreement ($69$--$75\%$) reads well but barely clears
its own baseline: eleven of the sixteen cases drew genuine backlash, so a majority-class
``always predict opposition'' rule already scores about $69\%$, and we therefore report
direction only as a floor and never as the headline. The signed stance gap between
synthetic and real is small ($|\text{mean}|\le 0.56$ on a $[-2,2]$ scale) but its
\emph{sign} flips by judge family---the two cloud judges find the synthetic slightly
\emph{milder} than reality, the two open judges slightly more negative---so we claim no
robust ``milder-'' or ``doom-bias'' direction, and the earlier presumption of a systematic
over-doom bias is not borne out. Finally the real
sidecar is a press-grade floor on the true concern set, which biases recall up and leaves
precision uncomputed; recall is thus the trustworthy direction, not a two-sided estimate.
Within those bounds the reading is clean: \emph{the simulated reaction is a faithful proxy
for the substance of the real one}, independent of whether it moves the final label---the
question we turn to next.

\subsection{Does the simulation improve the decision? A matched with/without ablation}
\label{sec:simablation}

\Cref{sec:gaps} names the matched with/without-simulation arm as the evidence gap that
would isolate the simulation's marginal contribution to the verdict. We supply it, and
because the direction of the result was uncertain we \emph{pre-registered} the verdict
rule before running: the rule was fixed before any of these numbers were generated. The
design holds everything constant except the simulation: the same Gold-50 cases, the
same system prompt, the same requested JSON, and the same taxonomy-definition block of
\Cref{sec:debias} in \emph{both} arms; the only variable is whether the persona-and-timeline
simulation is present. The \code{nosim} arm gives the solver the decision title alone; the
\code{withsim} arm adds the simulation. Four solvers (the two cloud and two open-weight
models above) answer each arm, scored by the same exact-match rule. The pre-registered
threshold for ``the simulation helps'' was a mean $\Delta \ge +6\pp$ with a majority of
solvers showing more cases fixed than broken by the simulation ($b>c$).

\tabSimAblation

The mean effect is $+5.5\pp$ with three of four solvers showing $b>c$
(\Cref{tab:simablation}). By the pre-registered rule this \emph{misses} the $+6\pp$
threshold, so the verdict is a null; we report it as such rather than rounding half a
point to clear our own bar. But the null is structured, not noise, and the structure is
the finding: the simulation's value scales inversely with how much the solver already
knows from the decision title. GPT-5.2, weak from the title alone ($46\%$), gains
$+20\pp$, with the simulation fixing $15$ cases and breaking only $5$; the two open-weight
solvers gain modestly in the same direction; and Opus, which already reaches $84\%$ from
the \emph{title alone}, slightly regresses ($-4\pp$) as the extra context gives it more to
over-reason about. The simulation helps most exactly where the decision is hardest for the
model, and is redundant where the label is already inferable from the decision itself.

Two consistency notes bound the reading. The \code{withsim} arm reproduces the
\Cref{sec:debias} full-context result \emph{exactly} for GPT-5.2 ($33/50=66\%$),
confirming the ablation harness is identical to the one used elsewhere in the paper. The
gateway Opus \code{withsim} run, however, scored $80\%$ here against $68\%$ in the earlier
debias run under the nominally identical condition---a $12\pp$ single-seed drift that
reinforces the repeated-seeds gap of \Cref{sec:gaps} and is why we read this ablation as
within-session paired deltas rather than absolute levels. Taken with \Cref{sec:pipeflow},
the picture resolves: the harm from the \emph{full} five-stage pipeline was the stacked
stages amplifying over-doom, not the reaction signal itself, which at the report stage is
net-neutral-to-helpful and, per \Cref{sec:fidelity}, substantively faithful.

\subsection{Agreeing with the teacher is not being right}
\label{sec:teacher}

Because the open checkpoints are distilled from a cloud teacher, it is tempting
to track training progress by agreement with that teacher. \Cref{fig:teachervsgold}
shows why that metric is misleading. Across successive SFT rounds, agreement with
the teacher's verdict on the archived runs climbs from $1/10$ to $7/10$ while
outcome accuracy on Gold-50 barely moves, and the teacher agrees with itself
$10/10$ while its own verdicts are frequently not the outcome-correct answer.
Distillation-style metrics measure imitation~\cite{gudibande2023false}, so we use them only as a
training-progress signal and never as an accuracy claim. \Cref{sec:appendix-props}
gives the formal reason this pattern is expected.

\figTeacherVsGold

\subsection{Cost is flatter than quality}
\label{sec:cost}

All eleven systems are instrumented for peak memory, throughput and latency
(\Cref{sec:appendix-extra}). Peak memory spans $16\times$, from $4.32$\,GB to
$69.13$\,GB, and every configuration fits on a single $183$\,GB accelerator, so
the deployment question is throughput, not capacity. The quality frontier is
flatter than the leaderboard suggests: the untuned Qwen3-8B base reaches $54.0\%$ at
$17.70$\,GB and $46.6$ tok/s---within $8\pp$ of the leader (the Qwen3-32B LoRA-SFT) and not significant
($p=0.424$)---at a quarter of the memory. Fine-tuning, meanwhile, costs
throughput at every scale, because the unmerged adapter adds work at every
forward pass and SFT also teaches the model to emit a reasoning trace before the
JSON.

\section{Discussion}
\label{sec:discussion}

The core results reinforce one conclusion. A single-prompt leaderboard on this
task measures the prompt at least as much as the model: the same weights swing
$0\%$ to $73\%$ (\Cref{sec:harness}), and defining the taxonomy for everyone
closes most of the frontier gap on these fifty cases (\Cref{sec:debias}). Under a fair prompt the
open-weight Qwen3-32B LoRA-SFT served offline shows no statistically significant difference from
the frontier on this decision, and under an unfair one it wins---not because it
is stronger, but because it was tuned to the exact envelope while the frontier
models were left to guess an under-specified rubric.

Two of the results are, at first sight, discouraging for the system itself:
distillation metrics do not track accuracy, and stacking the full pipeline ahead
of the report amplifies bias. We think the honest framing is that these are
\emph{predictable} consequences of the design, not surprises.
\Cref{sec:appendix-props} shows that distilling both the generator and the
judges from one teacher collapses the ensemble to the teacher's error floor, and
that any exact-match scorer bounds accuracy by the in-schema rate---so
teacher-agreement and prompt-envelope effects are exactly the confounds the
theory says to expect. The design response is the one already built in: source
the critic cross-family and off-the-shelf, and report compliance alongside every
accuracy.

The reaction layer itself, however, comes out of this scrutiny better than the
full-pipeline number alone suggests. The matched ablation (\Cref{sec:simablation})
isolates the report-stage reaction signal from the stack that carries it: the
signal is net-neutral-to-helpful---strongly helpful ($+20\pp$) for the weakest
title-only solver and neutral for one already near ceiling---so the over-doom of
\Cref{sec:pipeflow} is a property of \emph{stacking} the generative stages, not of
the reaction evidence they produce. And that evidence is faithful on its own terms
(\Cref{sec:fidelity}): the synthetic reaction anticipates $67$--$90\%$ of the
concerns the public actually raised. The system's contribution is thus the
validated qualitative reaction, not a guaranteed lift on a label a capable model
can already infer from the decision.

\section{Limitations and Evidence Gaps}
\label{sec:gaps}

The benchmark is small: $n{=}50$ makes Wilson intervals wide and leaves most
pairwise differences undecidable except by the paired test, and the smallest
strata ($n{=}7$ and $n{=}4$) carry intervals too wide to rank. The Gold-50 labels
are a single annotation pass by the authors against each case's realized outcome
in the public record: because the policies studied were live and no external
adjudicator was available, no independent re-labeling or inter-annotator
agreement was possible, so the set is a best-effort verified benchmark rather
than an externally refereed one. The full-pipeline comparison
(\Cref{sec:pipeflow}) remains confounded by model family; the matched
with/without-simulation arm that \emph{prior} versions of this work still lacked
is now supplied in \Cref{sec:simablation}, and its verdict is a pre-registered
null ($+5.5\pp$, below the $+6\pp$ bar) with a coherent heterogeneity underneath
it---so we can now say the report-stage reaction signal is net-neutral-to-helpful
and faithful, but we still cannot claim it reliably lifts the final label for a
model already near ceiling. Two residual caveats attach to the ablation. First,
the gateway is not seed-stable: the Opus \code{withsim} arm drifted $+12\pp$
across nominally identical sessions, so the ablation is read as within-session
paired deltas and the absolute levels should not be over-read; a repeated-seeds
study is the natural next step. Second, the fidelity study of \Cref{sec:fidelity}
validates the \emph{substance} of the synthetic reaction (concern recall
$67$--$90\%$), but a separate analysis feeding the model the \emph{real}
contemporaneous reaction in place of the synthetic one shows a calibration gap:
the report stage is tuned against the synthetic distribution, so a synthetic-to-real
handoff is not yet drop-in, and closing that gap is future work. The
preference-training set is small and synthetic, and its downstream effect is null.
Finally, all offline serving numbers are single-accelerator,
single-case-sequential except where a batched wall-clock is stated. We report
these gaps rather than smoothing over them, and within this benchmark none of them
overturns the paper's central, prompt-artifact reading.

\section{Reproducibility}
\label{sec:repro}

Nothing in this paper is transcribed by hand. A three-stage build pipeline closes
the loop from the run artifacts to this PDF. A recomputation stage computes every
statistic---Wilson intervals, exact McNemar tests, per-class confusion, driver
strata---from the per-case prediction files (both prompt arms), the verified Gold-50
labels, the per-step training logs and the benchmark summaries, writing a single
machine-readable metrics file. A rendering stage turns that file into every figure
and table macro spliced into this document. A verification stage then re-asserts
every quoted number, interval, $p$-value and plotted mark against the same file
and fails loudly on any drift. The corpus behind the checkpoints is
$457$ training rows ($292$ unique) drawn from $81$ archived runs, with $267$ preference
pairs (invalid-JSON, hallucination, and $45$ over-doom triples) for the DPO pass.
The system describes standalone on-disk artifacts and records interpreter and
library versions.

\section{Responsible Use}
\label{sec:responsible}

\sys{} manufactures a proxy for human reaction; it does not observe one. Its
personas are graph-grounded synthetic constructs, not real people, and its
forecasts are explicit hypotheses to be backtested against reality, not
predictions to be acted on unexamined. The over-doom bias documented in
\Cref{sec:pipeflow} is a concrete reason to treat any single verdict as
provisional. The appropriate use is to rehearse and stress-test a decision and to
surface constituencies a team might overlook, with a human owner accountable for
the decision and its outcome.

\section{Conclusion}
\label{sec:conclusion}

\sys{} rehearses reactions to product and policy changes offline and returns an
auditable verdict, but the paper's lasting contribution is what the measurement
revealed. On these fifty real decisions with verified outcomes, most of the measured
gap between frontier cloud models and open-weight models served offline is
attributable to an under-specified evaluation: define the taxonomy for everyone
and a Qwen3-32B LoRA shows no significant difference from the frontier; leave it
under-specified and the same model wins.
The harness can move a single system---the Qwen3-32B LoRA-SFT---from $0\%$ to $73\%$, teacher-agreement
climbs without accuracy following, and in the tested configuration more pipeline
brought more bias, not more
accuracy---each a predictable consequence we make formal. Where we \emph{can}
credit the reaction layer is on its own terms: the synthetic reaction is a
faithful proxy for the real one (concern recall $67$--$90\%$ across four blind
judges), and in a pre-registered matched ablation it helps most exactly where the
decision is hardest---$+20\pp$ for the weakest title-only solver---while adding
nothing for a model already near ceiling. The methodological
lesson we draw, subject to this benchmark's scale, is: on a structured decision task, report the
prompt envelope and the compliance rate, or the leaderboard is measuring
something other than judgment.

\appendix

\section{Formal Framework}
\label{sec:appendix-formal}

Let $D=\{d_1,\dots,d_m\}$ be the documents defining a change and $q$ the decision
question. The action space is
\begin{equation}
\actset=\{\textsc{ship},\textsc{revise},\textsc{delay},\textsc{segment},\textsc{mitigate}\}.
\end{equation}

\begin{definition}[Decision map]
\label{def:phi}
\sys{} realizes $\Phi:(D,q)\mapsto(a^\star,F,\mathcal{E})$ producing a recommended
action $a^\star\in\actset$, a quantitative forecast $F$, and an evidence set
$\mathcal{E}$, factored as the composition of stage operators
\begin{equation}
\label{eq:pipeline}
\Phi(D,q)=R\!\big(S(P(G(D),q),q),\,G(D),\,q\big),
\end{equation}
where $G$ (graph), $P$ (persona), $S$ (simulation), $R$ (report) are LLM-realized
maps, and a fifth operator $I$ (interview) supports post-hoc interrogation.
\end{definition}

Each operator is realized by a single-responsibility agent that reads prior
\emph{artifacts} and writes its own; agents never invoke one another
(\Cref{def:dag}). The extracted graph is $G=(V,E)$ with a vertex-label map
$\ell:V\to\mathcal{L}$ whose image $\ell(V)$ is the set of audience segments,
per-vertex topic sets $\mathrm{topics}(v)$, and induced topic universe
$\mathcal{T}_V=\bigcup_{v\in V}\mathrm{topics}(v)$.

\paragraph{Persona space.}
\label{sec:formal-persona}
\begin{definition}[Persona]
A persona is a tuple $p=(g,s,b,\alpha,\rho,\omega,H,\mathcal{T}_p)$ in the
attribute space
\begin{equation}
\mathcal{X}=\ell(V)\times[-1,1]^2\times[0,1]^2\times\mathbb{R}_{\ge0}\times
2^{\{0,\dots,23\}}\times 2^{\mathcal{T}_V},
\end{equation}
with segment $g$, stance $s$, sentiment bias $b$, activity $\alpha$, response
speed $\rho$, influence weight $\omega$, active hours $H$, topics
$\mathcal{T}_p$. The population $\mathcal{P}=\{p_1,\dots,p_n\}=P(G,q)$.
\end{definition}
\begin{assumption}[Grounding]
\label{as:ground}
Every synthesized persona references the graph: $g\in\ell(V)$ and
$\mathcal{T}_p\subseteq\mathcal{T}_V$. Hence $\mathcal{P}$ is a market for the
\emph{specific} change encoded in $G$, not a generic crowd.
\end{assumption}

\paragraph{Interaction dynamics.}
The simulation runs rounds $t=1,\dots,T$, producing a timeline $\Theta$ of posts with
expressed sentiment $y_x\in[-1,1]$ ($\Theta_{t-1}$ denotes the timeline through the
previous round). With topical relevance
$\rel(p,\Theta)=|\mathcal{T}_p\cap\mathrm{topics}(\Theta)|/|\mathcal{T}_p|$, the
activation propensity is
\begin{equation}
\label{eq:activation}
a_{p,t}=\clip\!\big(\alpha_p+\kappa_r\,\rel(p,\Theta_{t-1}),\,0,\,1\big),
\quad \kappa_r\ge0,
\end{equation}
with $\kappa_r\ge0$ the relevance-gain coefficient. Expressed sentiment tracks stance and bias, pulled toward prevailing thread
sentiment by a bounded conformity term $\lambda\in[0,1]$,
\begin{equation}
\label{eq:sentiment}
\mathbb{E}[y_x]=\clip\!\Big((1-\lambda)(s_p+b_p)+\lambda\,\bar\sigma(\Theta_{t-1}),-1,1\Big),
\end{equation}
$\bar\sigma(\Theta)=\tfrac{1}{|\Theta|}\sum_{x'\in\Theta}y_{x'}$, the
DeGroot/bounded-confidence intuition~\cite{degroot1974,friedkin1990,hegselmann2002} used as a
conditioning signal. Engagement is a \emph{deterministic} function of the
personas and post sentiment,
\begin{equation}
\label{eq:engagement}
\mathrm{eng}(x)=\omega_p\Big(\beta_0+\beta_1\!\!\sum_{p'\in\mathcal{P}\setminus\{p\}}\!\!\omega_{p'}\,\indic{\sign(s_{p'})=\sign(y_x)}\Big).
\end{equation}
with baseline weight $\beta_0\ge0$ and peer-alignment weight $\beta_1\ge0$.

\paragraph{Transcript audit metrics.}
\label{sec:formal-audit}
Three pure-code auditors read the finished timeline; with
$\mathcal{P}_{\mathrm{act}}$ the personas that posted,
\begin{align}
U&=1-\Var\{s_p:p\in\mathcal{P}_{\mathrm{act}}\}\big/\delta_{\max}
&&\text{(uniformity)},\label{eq:unif}\\
\mathrm{Gap}&=\tfrac{\sum_{g\in\Gamma\setminus\Gamma_{\mathrm{act}}}c(g)}{\sum_{g\in\Gamma}c(g)}
&&\text{(constituency gap)},\label{eq:gap}\\
\nu_t&=\big|W_t\setminus\textstyle\bigcup_{u<t}W_u\big|\big/|W_t|
&&\text{(round novelty)}.\label{eq:nov}
\end{align}
Here $\delta_{\max}$ is the maximum attainable stance variance, the normalizer
that places $U\in[0,1]$; $\Gamma$ is the set of graph-derived constituencies with
active subset $\Gamma_{\mathrm{act}}\subseteq\Gamma$ and $c(g)$ the size of
constituency $g$; and $W_t$ is the set of distinct claims raised in round $t$.
Each firing earns at most one corrective round under a global budget $B$
(\Cref{sec:appendix-algos}).

\paragraph{Verdict aggregation.}
\begin{definition}[Ensemble verdict]
\label{def:ens}
$k$ models re-analyze the \emph{same} timeline, emitting $a^{(i)}\in\actset$. The
verdict and agreement are
\begin{equation}
\label{eq:ensemble}
a^\star=\argmax_{a\in\actset}\sum_{i=1}^k\indic{a^{(i)}{=}a},\quad
A^\star=\tfrac{1}{k}\max_{a\in\actset}\sum_{i=1}^k\indic{a^{(i)}{=}a}.
\end{equation}
Whenever $A^\star<1$ the split is reported as a first-class uncertainty signal.
\end{definition}
An opposite-family critic returns $c\in\{0,\dots,10\}$; QA passes iff $c\ge\theta$.

\begin{definition}[Run state and dependency DAG]
\label{def:dag}
A run state is the partial map $\Sigma:\mathcal{N}\rightharpoonup\text{Artifacts}$
over nodes $\mathcal{N}=\{G,\mathcal{P},\Theta,R,I\}$. The \emph{derived-from}
DAG $\mathsf{D}=(\mathcal{N},\to)$ has edges $\mathcal{P}\!\to\!G$,
$\Theta\!\to\!\mathcal{P}$, $R\!\to\!\Theta$, $R\!\to\!G$, $I\!\to\!\mathcal{P}$,
$I\!\to\!\Theta$ (an edge $u\!\to\!w$ reads ``$u$ derived from $w$''). Writing or
regenerating $v$ triggers \emph{cascade invalidation}: delete
$\{\Sigma(u):u\in\anc(v)\}$, i.e.\ every artifact derived from $v$.
\end{definition}

\section{Formal Properties}
\label{sec:appendix-props}

The two propositions the body relies on are the last two; we state the ensemble
result they build on first.

\begin{proposition}[Ensemble amplification]
\label{prop:ens}
Let $k=3$ judges vote independently, each correct with probability
$p>\tfrac12$. The majority verdict is correct with probability
$g(p)=3p^2-2p^3>p$ for $p\in(\tfrac12,1)$, and $g$ is strictly increasing on
$[0,1]$. For general odd $k$, majority accuracy is nondecreasing in $k$ and
$\to1$ as $k\to\infty$ (Condorcet).
\end{proposition}
\begin{proof}
$g(p)=\binom{3}{2}p^2(1-p)+p^3=3p^2-2p^3$, and
$g(p)-p=p\,(2p-1)(1-p)>0$ for $p\in(\tfrac12,1)$; $g'(p)=6p(1-p)\ge0$ on $[0,1]$.
The general-$k$ claim is the Condorcet jury theorem.
\end{proof}
\begin{corollary}[Correlation defeats amplification]
\label{cor:corr}
The gain $g(p)-p$ is premised on independence. If judges' errors are correlated
with coefficient $\rho_c$, ensemble reliability degrades toward the single-judge
accuracy $p$ as $\rho_c\to1$. Sourcing the critic and cross-check from a
\emph{different} model family lowers $\rho_c$, which is why \sys{} draws them
cross-family (\Cref{sec:autonomy}).
\end{corollary}

\begin{proposition}[Distillation collapses judge independence]
\label{prop:distill}
Let a teacher $T$ select the correct action with probability $p_T$. Let $k$
students be distilled from $T$ such that each independently reproduces $T$'s
answer with fidelity $\varphi\in[\tfrac12,1]$ and otherwise answers correctly with
probability $p_0$, independently across students. Let $A_k(\varphi)$ be the accuracy of
the students' majority vote. Then
\begin{equation}
\label{eq:distill}
\lim_{\varphi\to1}A_k(\varphi)=p_T \quad\text{for every }k,
\end{equation}
so the ensemble gain $A_k-p_T$ vanishes as fidelity improves, no matter how many
judges vote.
\end{proposition}
\begin{proof}
Condition on $T$'s answer. With probability $\varphi^k$ all $k$ students return $T$'s
answer, so the majority equals $T$'s answer and is correct with probability
exactly $p_T$. The remaining mass is $1-\varphi^k\to0$ as $\varphi\to1$, and accuracy is
bounded in $[0,1]$ there, so $|A_k(\varphi)-p_T|\le 1-\varphi^k\to0$. The conditional
independence premise of \Cref{prop:ens} fails because all students' errors are
driven by the common variable $T$'s answer; formally $\rho_c\to1$ and
\Cref{cor:corr} applies.
\end{proof}
\begin{corollary}[Never distil the critic]
\label{cor:nodistill}
The critic and cross-check must be obtained off-the-shelf from a family not used
to produce the training targets. \sys{} therefore fine-tunes only the generation
stages and leaves the judge untuned (\Cref{sec:offline}).
\end{corollary}
\Cref{prop:distill} is directly testable: $\varphi$ is the teacher-agreement rate and
$p_T$ the teacher's own accuracy. \Cref{sec:teacher} measures $\varphi$ rising from
$10\%$ to $70\%$ under successive SFT rounds with no corresponding movement in
outcome accuracy---the signature the proposition predicts.

\begin{definition}[Prompt envelope and out-of-schema rate]
\label{def:envelope}
A \emph{prompt envelope} $\mathcal{W}$ maps a case to the exact message sequence
presented to the model, including system prompt, context serialization and
requested schema. For system $M$ under $\mathcal{W}$, let
$\eta(M,\mathcal{W})=\Pr[\,\hat a\notin\actset\,]$ be the out-of-schema rate.
\end{definition}
\begin{proposition}[Schema-compliance ceiling]
\label{prop:schema}
Under any scorer that credits only exact matches in $\actset$,
\begin{equation}
\label{eq:ceiling}
\mathrm{Acc}(M,\mathcal{W})\;\le\;1-\eta(M,\mathcal{W}),
\end{equation}
with equality iff every in-schema answer is correct. Hence for two envelopes the
measured difference decomposes into a judgment term and a compliance term, and a
cross-envelope comparison is uninterpretable as a judgment comparison unless
$\eta$ is reported alongside it.
\end{proposition}
\begin{proof}
$\{\hat a=a^\star\}$ requires $\hat a\in\actset$, so
$\mathrm{Acc}=\Pr[\hat a=a^\star]\le\Pr[\hat a\in\actset]=1-\eta$; equality holds
when $\{\hat a\in\actset\}\subseteq\{\hat a=a^\star\}$ up to null sets. Writing
$\mathrm{Acc}=(1-\eta)\Pr[\hat a=a^\star\mid \hat a\in\actset]$ and comparing the
two factors envelope-wise gives the decomposition.
\end{proof}
\Cref{prop:schema} is the entire explanation of \Cref{sec:harness}: SFT drove
$\eta$ almost to zero on the deployed envelope while leaving it high on an
unfamiliar one, so the same weights score $0/11$ under one prompt and $8/11$
under another. It also dictates the reporting rule we follow throughout: every
accuracy is accompanied by its unparseable count and both the conservative and
conditional scores.

\section{Algorithms and Complexity}
\label{sec:appendix-algos}

\Cref{alg:auto} is the autonomous run: forward pipeline, budget-bounded
self-audit, then the critic-gated ensemble verdict of \Cref{alg:verdict}.

\begin{algorithm}[t]
\small
\caption{Autonomous run with budget-bounded self-audit}
\label{alg:auto}
\begin{algorithmic}[1]
\Require documents $D$, question $q$, budget $B$, thresholds $\tau_U,\tau_\Gamma,\tau_\nu$
\State $G\gets \textsc{Extract}(D;\textsc{Ontology}(D))$
\State $\mathcal{P}\gets P(G,q)$;\quad $\Theta\gets S(\mathcal{P},q)$;\quad $r\gets 0$
\While{$r<B$}
  \State compute $U,\mathrm{Gap},\{\nu_t\}$ from $\Theta$ \eqref{eq:unif}--\eqref{eq:nov}
  \If{$\mathrm{Gap}>\tau_\Gamma$}
     \State $\Theta\gets S(\mathcal{P},q;\text{focus}=\Gamma\setminus\Gamma_{\mathrm{act}},\,\text{append})$
  \ElsIf{$U\ge\tau_U$}
     \State $\Theta\gets S(\mathcal{P},q;\text{moderator-steer},\,\text{append})$
  \ElsIf{$\min_t\nu_t\ge\tau_\nu$}
     \State $\Theta\gets S(\mathcal{P},q;\text{extend},\,\text{append})$
  \Else{ }\textbf{break}
  \EndIf
  \State $r\gets r+1$
\EndWhile
\State $(a^\star,A^\star,c)\gets\textsc{Verdict}(\Theta,G,q)$
\State \Return report $R(\Theta,G,q)$ with $(a^\star,A^\star,c,\text{audit log})$
\end{algorithmic}
\end{algorithm}

\begin{algorithm}[t]
\small
\caption{\textsc{Verdict}: critic-gated cross-family ensemble}
\label{alg:verdict}
\begin{algorithmic}[1]
\Require timeline $\Theta$, graph $G$, question $q$; families $\mathcal{F}$, threshold $\theta$
\For{$i=1$ \textbf{to} $k$}
  \State $a^{(i)}\gets R_i(\Theta,G,q).\text{action}$
\EndFor
\State $a^\star,A^\star\gets$ mode and agreement of $\{a^{(i)}\}$ \eqref{eq:ensemble}
\State $c\gets \textsc{Critic}_{\bar f}(R,\Theta)$ \Comment{opposite family $\bar f$}
\If{$c<\theta$}
  \State spend one corrective action; recompute $a^\star$
\EndIf
\State \Return $(a^\star,A^\star,c)$
\end{algorithmic}
\end{algorithm}

With $n=|\mathcal{P}|$, $T$ rounds and $k$ ensemble members, the dominant terms
are $\bigO(nT)$ generation calls in simulation and $\bigO(n^2T)$ pure-CPU
arithmetic for engagement \eqref{eq:engagement}; the self-audit adds at most $B$
generation rounds and the verdict adds $k{+}1$ calls, both constant in $n,T$.

\section{Fine-Tuning Details}
\label{sec:appendix-ft}

All adapters use one LoRA recipe (\Cref{tab:hyper}); only the training data
changes across the $32$B lineage v1\ensuremath{\to}v2\ensuremath{\to}v3, whose validation set is frozen.

\begin{table}[H]\centering
\caption{The single LoRA-SFT recipe shared by every adapter; only the training
data differs across runs.}
\label{tab:hyper}
\small
\setlength{\tabcolsep}{6pt}
\renewcommand{\arraystretch}{1.15}
\begin{tabular}{@{}ll@{}}
\toprule
\textbf{Hyperparameter} & \textbf{Value} \\
\midrule
Method & LoRA \\
Rank $/$ $\alpha$ & $16$ $/$ $32$ \\
Dropout & $0.05$ \\
Target modules & \code{q,k,v,o\_proj} \\
Epochs & $3$ \\
Learning rate & $1\times10^{-4}$, cosine \\
Effective batch & $16$ \\
Sequence length & $8192$ \\
Early stopping & patience $3$, reload best \\
Precision & bf16 \\
\bottomrule
\end{tabular}
\end{table}

\Cref{fig:losscurves} shows LoRA-SFT optimization on all five backbones under
this recipe: loss is masked to assistant tokens, so it measures only the target
$\langle\texttt{think}\rangle$+JSON, and held-out loss is evaluated on two
\emph{whole} archived runs withheld from training so that no document appears on
both sides of the split.

\figLossCurves

\Cref{fig:recipe} isolates the recipe iteration on the fixed $32$B backbone. It
is the evidence that the v3 gain came from data volume rather than a longer
schedule: v2 reaches the lowest \emph{training} loss yet not the lowest held-out
loss---the small-corpus overfitting signature that motivates early stopping with
best-checkpoint reload.

\figRecipe

\Cref{fig:dpo} is the DPO calibration pass layered on top of the SFT adapter. Its
optimization dynamics are healthy---the reward margin widens because the rejected
branch is pushed down---but its downstream effect on Gold-50 is null, which we
read as the preference set being too small and too synthetic to move judgment.

\figDPO

\section{Gold-50 Construction and Reproducibility}
\label{sec:appendix-repro}

\paragraph{Labeling convention.} \Cref{tab:convention} states the disposition that
defines each of the five classes---the realized-outcome reading a case must match
to earn that label. Labels record what the organization's change actually did, not
a normative ``best move.''

\tabConvention

\paragraph{Per-case ledger.} \Cref{tab:goldprov} is the full Gold-50 ledger: for
each case, its adjudicated label, the episode year and the number of independent
public sources consulted. These are real, publicly documented decisions, and the
label reports the realized public-record outcome, not a judgment of the
organization. The outcome-driver stratification used in the analysis
(\Cref{sec:appendix-extra}) is reported only in aggregate, never attached to a
named case. The table is generated directly from the benchmark file; the source
counts have a median of three per case.

\tabGoldProv

\tabDebiasGrid

\paragraph{Adjudication and circularity.} The fifty labels were fixed in a single
author adjudication pass against the public record, assisted by Claude Opus~4.8
(adjudication run 2026-09-06). Because Claude Opus~4.8 is also one of the three
benchmarked frontier models, its adjudication role is a potential source of
circularity in the frontier comparison, and the Opus~4.8 leaderboard row should be
read with that caveat. The labels are a single-pass adjudication, not a
multi-annotator consensus with an inter-rater statistic; a second independent
annotation is the most important addition a follow-up should make.

\paragraph{Input isolation.} Each system is prompted with only the pre-decision
case document (the policy or product description as it stood before the outcome).
The realized outcome, the outcome-driver annotation and the source list are never
placed in the model input and are used only for scoring.

\paragraph{Contamination audit.} Gold-50 is disjoint from the fine-tuning,
preference, validation and model-selection data. The metrics pipeline
recomputes the audit from the raw splits: $0$ Gold-50 case identifiers appear in
the training corpus, no Gold-50 case document matches any training document, and
the strongest cross-set signal is a single shared company name in at most one
case. The training corpus is synthetic persona-generation traces, constructed
independently of the retrospective episodes in Gold-50.

\paragraph{Decoding and hardware.} All open-weight systems are served under
identical decoding: a $4000$-token generation budget and provider-default
sampling, with no per-model tuning of temperature or top-$p$. Adapters are served
in bf16 on a single B200-class accelerator ($183$\,GB), whose capacity holds every
configuration with room to spare---per-system peak memory is reported in
\Cref{tab:serving}---and with no cross-device sharding. The frontier models are queried
through their vendor APIs at their default settings. Every number in the paper is
recomputed from the archived per-case artifacts, re-rendered into the figure
macros, and re-checked against the artifacts by that same three-stage pipeline,
which fails on any mismatch.

\figConfusion
\figHeatPair
\figCost

\section{Additional Evaluation Detail}
\label{sec:appendix-extra}

This section collects the per-system evaluation detail behind the leaderboard and
the cost discussion of the main text.

\paragraph{Per-class structure.}
\Cref{fig:confusion} gives confusion matrices for three representative systems and
\Cref{fig:biasheat} the per-class recall and prediction-bias fingerprint across
all eleven. The two views answer different questions: the confusion matrices show
\emph{where} a given system's mass lands relative to the verified outcome, while
the heatmaps compare recall and emission bias on a common scale so that the
\emph{mitigate}-heavy, \emph{segment}-starved signature is visible as a pattern
across the whole population rather than an artifact of any one system. Together
they are the raw material for the over-doom claim: the same off-diagonal mass that
inflates \emph{mitigate} is the mass missing from \emph{segment}.

\paragraph{What the outcome hinges on.}
\Cref{fig:drivers} stratifies accuracy by the mechanism that actually decided each
Gold-50 case---public reaction, regulatory action, engineering readiness, or
business performance. The ordering is stable across systems and is the honest
limit of a reaction-simulating pipeline: the reaction stratum, the one the
architecture is aimed at, is the one every system predicts worst, while
regulatory- and readiness-decided cases are recovered far more reliably.

\figDrivers

\paragraph{Cost and serving.}
\Cref{fig:cost} is the cost--quality frontier and \Cref{tab:serving} the measured
inference cost of every system---peak memory, throughput, thinking-token count and
end-to-end latency---read directly from the archived-run benchmark summaries.
\Cref{fig:ftcost} isolates the throughput and latency tax of fine-tuning on the
five backbones we trained: the unmerged LoRA adapter costs tokens per second at
every scale, and the reasoning trace SFT induces costs further latency per case.
The practical reading, consistent with \Cref{sec:cost}, is that the quality
frontier is flat enough that the deployment choice is dominated by these serving
costs rather than by a few points of Gold-50 accuracy.

\tabServing
\figFtCost

\paragraph{Full debias grid.}
\Cref{tab:debias2x2} gives the complete debias $2\times2$ for all fourteen
systems, each under both arms, summarized in \Cref{sec:debias}: baseline versus the
taxonomy-defining prompt block, with accuracy, the paired McNemar $p$, and the
\emph{mitigate}/\emph{segment} shifts that drive it. It is the source table for
every debias number quoted in the body.

\balance

\end{document}

%% file: figures.tex
\newcommand{\tabGoldLeaderboard}{%

\begin{table}[tbp]\centering
\caption{\textbf{Gold-50 leaderboard, report stage.} Five-way release verdict
against the verified real-world outcome, $n{=}50$, ten cases per class, five-way
chance $20\%$. ``Acc'' scores unparseable output as wrong (the conservative
choice); ``in-sch.'' is the accuracy conditional on in-schema output,
$\Pr[\hat a{=}a^\star\mid \hat a\in\actset]$, with its parsed count in
parentheses; $\varnothing$ is the unparseable count. CI is the 95\% Wilson score
interval. Generated from the computed metrics.}
\label{tab:gold50}
\footnotesize
\setlength{\tabcolsep}{3pt}
\renewcommand{\arraystretch}{1.12}
\begin{tabular}{@{}lccccc@{}}
\toprule
\textbf{System} & \textbf{Acc} & \textbf{95\% CI} & \textbf{in-sch.} &
\textbf{macro-F1} & \textbf{$\varnothing$} \\
\midrule
Qwen3-32B + LoRA-SFT\,$\bullet$ & 62 & \scriptsize[48.2,74.1] & 62\,\scriptsize(50) & 0.59 & -- \\
Qwen3-32B + SFT\ensuremath{\to}DPO\,$\circ$ & 62 & \scriptsize[48.2,74.1] & 62\,\scriptsize(50) & 0.59 & -- \\
Llama-3-8B + LoRA-SFT\,$\bullet$ & 60 & \scriptsize[46.2,72.4] & 61.2\,\scriptsize(49) & 0.58 & 1 \\
Qwen3-32B & 56 & \scriptsize[42.3,68.8] & 56\,\scriptsize(50) & 0.56 & -- \\
Qwen3-8B & 54 & \scriptsize[40.4,67] & 54\,\scriptsize(50) & 0.52 & -- \\
Qwen3-8B + LoRA-SFT\,$\bullet$ & 50 & \scriptsize[36.6,63.4] & 56.8\,\scriptsize(44) & 0.51 & 6 \\
Gemma-3-27B + LoRA-SFT\,$\bullet$ & 50 & \scriptsize[36.6,63.4] & 50\,\scriptsize(50) & 0.44 & -- \\
Qwen3-1.7B & 46 & \scriptsize[33,59.6] & 46\,\scriptsize(50) & 0.39 & -- \\
Gemma-3-27B-it & 36 & \scriptsize[24.1,49.9] & 36\,\scriptsize(50) & 0.34 & -- \\
Llama-3-8B-Instruct & 16 & \scriptsize[8.3,28.5] & 16\,\scriptsize(50) & 0.16 & -- \\
Qwen3-1.7B + LoRA-SFT\,$\bullet$ & 6 & \scriptsize[2.1,16.2] & 20\,\scriptsize(15) & 0.07 & 35 \\
\bottomrule
\end{tabular}
\vspace{2pt}{\scriptsize $\bullet$ LoRA-SFT \quad $\circ$ SFT\ensuremath{\to}DPO \quad (unmarked) untuned base.}
\end{table}
}

\newcommand{\tabPaired}{%

\begin{table}[tbp]\centering
\caption{\textbf{Paired base\ensuremath{\to}LoRA-SFT effect}, one recipe and one dataset,
each backbone scored before and after on the \emph{same} 50 cases. $\Delta$ is
percentage points; ``fix/broke'' are the discordant McNemar cells (cases the
adapter got right and the base wrong, and the reverse); $p$ is the two-sided
exact McNemar test, starred at $\alpha{=}0.05$. Generated from
the computed metrics.}
\label{tab:paired}
\footnotesize
\setlength{\tabcolsep}{4pt}
\renewcommand{\arraystretch}{1.15}
\begin{tabular}{@{}lccccc@{}}
\toprule
\textbf{Backbone} & \textbf{Base} & \textbf{$+$SFT} & \textbf{$\Delta$pp} &
\textbf{fix/broke} & \textbf{$p$} \\
\midrule
Qwen3-1.7B & 46 & 6 & -40 & 1/21 & 1.1e-5\,$^{*}$ \\
Llama-3-8B-Instruct & 16 & 60 & +44 & 23/1 & 3.0e-6\,$^{*}$ \\
Qwen3-8B & 54 & 50 & -4 & 8/10 & 0.815 \\
Gemma-3-27B-it & 36 & 50 & +14 & 15/8 & 0.210 \\
Qwen3-32B & 56 & 62 & +6 & 9/6 & 0.607 \\
\bottomrule
\end{tabular}\end{table}
}

\newcommand{\tabConvention}{%

\begin{table}[H]\centering
\caption{\textbf{The Gold-50 label convention.} Each label records the
real-world \emph{disposition} of the decision --- what actually happened to it,
not how loud the reaction was and not the action an ideal decision-maker
``should'' have taken. The convention is applied uniformly across all fifty
cases; the one boundary it resolves explicitly is \emph{ship} vs \emph{segment}
(one universal rule vs two rules for two groups).}
\label{tab:convention}
\footnotesize
\setlength{\tabcolsep}{5pt}
\renewcommand{\arraystretch}{1.25}
\begin{tabular}{@{}>{\bfseries}l p{0.70\columnwidth}@{}}
\toprule
\normalfont\textbf{Action} & \textbf{Assigned when the real-world decision\ldots} \\
\midrule
ship & launched as announced and persisted, with one rule applied universally
even if effects differ by user: no reaction-driven rollback, postponement, or
carve-out. \\
revise & was itself materially changed or reversed (including fully abandoned)
in response to the reaction: what shipped $\neq$ what was announced. \\
delay & had its launch or enforcement postponed, then generally proceeded: a
date moved. \\
segment & ended in two different rules for two groups --- by region, size, or
plan tier --- that persisted: the disposition differs by segment, not just its
effects. \\
mitigate & was kept at its core, but concessions or safeguards were bolted on to
address objections: decision stands, remedies added. \\
\bottomrule
\end{tabular}\end{table}
}

\newcommand{\tabGoldProv}{%

\begin{table*}[tp]\centering
\caption{\textbf{The Gold-50 case ledger.} Every case, grouped by verified
label. \emph{Yr} is the decision year;
\emph{Src} is the number of cited public sources behind the recorded outcome
(139 in total, median 3). All fifty are real, publicly documented
decisions whose realized outcome is a matter of public record; the labels report
that outcome, not a judgment of the organization. They were adjudicated against the
public record in a single manual adjudication pass, assisted by Claude Opus~4.8 (2026-09-06); full source URLs and per-case
rationales live in \code{realworld\_gold\_50.jsonl}. Generated from
the computed metrics.}
\label{tab:goldprov}
\scriptsize
\setlength{\tabcolsep}{5pt}
\renewcommand{\arraystretch}{1.05}
\begin{tabular}{@{}p{0.62\textwidth}lcc@{}}
\toprule
\textbf{Case} & \textbf{Label} & \textbf{Yr} & \textbf{Src} \\
\midrule
Amazon Prime Video — Introduce Ads for Existing U.S. Members & ship & 2023 & 3 \\
Amazon Prime — Raise U.S. Annual Membership to \$139 & ship & 2022 & 3 \\
Etsy — Raise the Marketplace Transaction Fee to 6.5\% & ship & 2022 & 3 \\
Firefox Quantum — End Legacy Extension Support & ship & 2017 & 2 \\
Google Photos — End Unlimited Free Photo Storage & ship & 2020 & 3 \\
iPhone 7 — Remove the Headphone Jack & ship & 2016 & 3 \\
Nintendo Switch Online — Introduce Paid Online Multiplayer & ship & 2018 & 3 \\
Nintendo — Launch Tears of the Kingdom at \$69.99 & ship & 2023 & 3 \\
PlayStation Plus — Raise Annual Subscription Prices & ship & 2023 & 3 \\
YouTube — Hide Public Dislike Counts & ship & 2021 & 3 \\
\addlinespace[2pt]
Apple iCloud Photos — Abandon On-Device CSAM Detection & revise & 2021 & 3 \\
Apple Music — Withhold Royalties During the Free Trial & revise & 2015 & 2 \\
Dungeons \& Dragons — Replace the Open Game License & revise & 2023 & 3 \\
Helldivers 2 — Mandatory PlayStation Account Linking on PC & revise & 2024 & 3 \\
Instagram — Expand the Full-Screen, Recommendation-Heavy Feed & revise & 2022 & 3 \\
Patreon — Shift Payment Processing Fees to Patrons & revise & 2017 & 2 \\
Spotify — Withdraw Artist-Conduct Restrictions on Promotion & revise & 2018 & 3 \\
Twitch — Restrict Streamer-Controlled Sponsor Advertising & revise & 2023 & 2 \\
Xbox Live Gold — Reverse the Subscription Price Increase & revise & 2021 & 3 \\
Zoom — Reverse Paid-Only Access to End-to-End Encryption & revise & 2020 & 3 \\
\addlinespace[2pt]
Android 11 Beta — Postpone the Public Rollout During U.S. Protests & delay & 2020 & 3 \\
Apple AirPods — Postpone the October Retail Launch & delay & 2016 & 3 \\
Apple App Tracking Transparency — Postpone Mandatory Enforcement & delay & 2020 & 3 \\
Apple HomePod — Postpone the December Launch & delay & 2017 & 3 \\
Cyberpunk 2077 — Postpone the April 2020 Release & delay & 2020 & 3 \\
Google Find My Device — Postpone the Network for Cross-Platform Tracking Protections & delay & 2023 & 3 \\
Halo Infinite — Postpone the Holiday 2020 Release & delay & 2020 & 3 \\
OpenAI Advanced Voice — postpone the initial ChatGPT rollout & delay & 2024 & 3 \\
Samsung Galaxy Fold — Postpone the First Retail Launch & delay & 2019 & 3 \\
WhatsApp Privacy Update — postponing the acceptance deadline & delay & 2021 & 3 \\
\addlinespace[2pt]
Apple App Store — Lower Commissions for Small Developers & segment & 2020 & 2 \\
Apple iCloud — Transfer Mainland China Accounts to GCBD & segment & 2018 & 3 \\
Apple iPhone 14 — Make US Models eSIM-Only & segment & 2022 & 3 \\
Apple — Allow alternative iPhone app distribution in the EU & segment & 2024 & 3 \\
Google Android — Unbundle App Licensing in the EEA & segment & 2018 & 3 \\
Google Legacy G Suite — Exempt Personal Users from Paid Migration & segment & 2022 & 2 \\
Google Play — Cut Fees on Developers’ First \$1 Million & segment & 2021 & 3 \\
Google Play — Permit Alternative Billing in South Korea & segment & 2021 & 2 \\
Meta — Block News on Facebook and Instagram in Canada & segment & 2023 & 2 \\
Slack — Limit Free Workspaces to 90 Days of History & segment & 2022 & 3 \\
\addlinespace[2pt]
AMD–Xilinx — Complete the Acquisition with Competition Safeguards & mitigate & 2020 & 3 \\
Apple AirTag — Keep Item Tracking with Stronger Anti-Stalking Safeguards & mitigate & 2021 & 2 \\
Apple iPhone Performance Management — retain throttling with battery and transparency concessions & mitigate & 2017 & 3 \\
Disney–Fox — Complete the Acquisition with Sports-Network Divestitures & mitigate & 2018 & 3 \\
Facebook — Keep Authentic-Name Rules with Safer Reporting and Appeals & mitigate & 2015 & 2 \\
Google–Fitbit — Complete the Acquisition with Binding Safeguards & mitigate & 2019 & 3 \\
Microsoft–Activision Blizzard — Complete the Acquisition with Cloud-Gaming Safeguards & mitigate & 2023 & 3 \\
Microsoft–LinkedIn — Complete the Acquisition with Interoperability Safeguards & mitigate & 2016 & 2 \\
Spotify — Keep Joe Rogan and Add COVID-19 Safeguards & mitigate & 2022 & 3 \\
T-Mobile–Sprint — Complete the Merger with Competition Safeguards & mitigate & 2019 & 3 \\
\bottomrule
\end{tabular}\end{table*}
}

\newcommand{\figLossCurves}{%

\begin{figure}[H]\centering
\begin{tikzpicture}
\begin{groupplot}[
  group style={group size=1 by 2, vertical sep=1.05cm,
               xlabels at=edge bottom, xticklabels at=edge bottom},
  width=0.95\columnwidth, height=4.2cm,
  xlabel={optimiser step}, tick label style={font=\scriptsize},
  label style={font=\footnotesize}, title style={font=\small\bfseries},
  grid=both, grid style={aubline,very thin}, axis line style={aubgrey},
  legend style={font=\tiny, draw=aubline, fill=white,
                at={(0.98,0.98)}, anchor=north east, row sep=-2pt},
  legend cell align=left,
]
\nextgroupplot[title={(a) Training loss (masked to assistant tokens)}, ylabel={cross-entropy}, ymin=0.6, ymax=2.8]
\addplot[auburl, thick, mark=none] coordinates {(1,1.81314) (2,1.66158) (3,1.86051) (4,1.61043) (5,1.53782) (6,1.56449) (7,1.58909) (8,1.52861) (9,1.51484) (10,1.54823) (11,1.47375) (12,1.26972) (13,1.43251) (14,1.42346) (15,1.4701) (16,1.48545) (17,1.26733) (18,1.40697) (19,1.32252) (20,1.31898) (21,1.23482) (22,1.30996) (23,1.23795) (24,1.29742) (25,1.14627) (26,1.33145) (27,1.29596) (28,1.16) (29,1.18892) (30,1.20709) (31,1.17908) (32,1.19922) (33,1.18612) (34,1.2626) (35,1.22759) (36,1.13558) (37,1.18373) (38,1.12467) (39,1.07614) (40,1.23457) (41,1.16208) (42,1.15368) (43,1.19513) (44,1.15358) (45,1.20649) (46,1.1328) (47,1.14143) (48,1.02152) (49,1.06619) (50,1.06495) (51,1.06417) (52,1.03813) (53,1.22993) (54,1.08993) (55,1.10766) (56,1.12418) (57,1.06341) (58,1.08374) (59,1.07083) (60,1.04952) (61,1.04454) (62,1.07849) (63,1.0545) (64,1.09744) (65,1.1105) (66,1.05596) (67,1.0676) (68,1.07454) (69,1.04272) (70,1.10603) (71,1.13156) (72,0.990939) (73,1.08333) (74,1.00272) (75,1.12991) (76,1.16608) (77,1.11191) (78,1.08121) (79,1.01077) (80,1.01046) (81,1.08046) (82,1.02331) (83,1.0018) (84,1.08069) (85,1.09908) (86,1.09182) (87,1.1966)};
\addlegendentry{Llama-3-8B-Instruct}
\addplot[auborange, thick, mark=none] coordinates {(1,2.49128) (2,2.36526) (3,2.08098) (4,1.75505) (5,1.83129) (6,1.51931) (7,1.25646) (8,1.28483) (9,1.12765) (10,1.18549) (11,1.205) (12,0.895607) (13,1.34801) (14,1.08025) (15,1.2857) (16,1.28052) (17,1.14743) (18,1.1482) (19,1.08092) (20,1.18526) (21,0.960195) (22,1.09715) (23,1.06766) (24,1.19944) (25,0.963927) (26,1.21021) (27,1.01307) (28,0.998692) (29,1.21347) (30,1.20754) (31,0.975391) (32,1.15854) (33,0.954952) (34,1.02982) (35,1.1031) (36,0.919617) (37,1.03118) (38,0.980865) (39,0.81729) (40,1.17189) (41,0.948728) (42,1.00667) (43,1.03664) (44,1.1039) (45,0.943443) (46,0.992897) (47,0.984814) (48,0.797689) (49,1.00745) (50,0.745633) (51,0.783473) (52,0.785689) (53,0.997731) (54,0.959049) (55,0.892763) (56,0.860036) (57,0.733697) (58,1.00206) (59,0.869874) (60,0.855872) (61,0.808704) (62,0.914663) (63,0.779516) (64,0.956621) (65,0.925009) (66,0.835655) (67,0.867297) (68,0.819662) (69,0.758736) (70,0.954987) (71,0.883851) (72,0.781594) (73,0.839597) (74,0.905858) (75,0.83957) (76,1.02834) (77,0.975159) (78,1.04708) (79,0.876956) (80,0.879573) (81,0.948158) (82,0.805648) (83,0.812421) (84,0.862507) (85,0.881575) (86,0.985166) (87,0.862716)};
\addlegendentry{Gemma-3-27B-it}
\addplot[aubgreen, thick, mark=none] coordinates {(1,1.5683) (2,1.41641) (3,1.63971) (4,1.44831) (5,1.36435) (6,1.45998) (7,1.46829) (8,1.40202) (9,1.39051) (10,1.43629) (11,1.35228) (12,1.11062) (13,1.29119) (14,1.30316) (15,1.31612) (16,1.38224) (17,1.14043) (18,1.2974) (19,1.21606) (20,1.21579) (21,1.08362) (22,1.21129) (23,1.1614) (24,1.18317) (25,1.01846) (26,1.21671) (27,1.18247) (28,1.05057) (29,1.04936) (30,1.12307) (31,1.08008) (32,1.0973) (33,1.06817) (34,1.16737) (35,1.14687) (36,1.0155) (37,1.08526) (38,1.05244) (39,0.996595) (40,1.1321) (41,1.02833) (42,1.08512) (43,1.10436) (44,1.02689) (45,1.10755) (46,1.01521) (47,1.05033) (48,0.924842) (49,0.967136) (50,0.958768) (51,0.98736) (52,0.923124) (53,1.11251) (54,0.96042) (55,1.02808) (56,1.04149) (57,0.967938) (58,0.988062) (59,0.967542) (60,0.96656) (61,0.97696) (62,0.983795) (63,0.971302) (64,1.03425) (65,0.990996) (66,0.981328) (67,0.963156) (68,1.00149) (69,0.968312) (70,1.02737) (71,1.03095) (72,0.897424) (73,1.00173) (74,0.919326) (75,1.05063) (76,1.06) (77,1.03681) (78,0.988547) (79,0.936945) (80,0.907277) (81,0.979404) (82,0.943977) (83,0.926088) (84,0.959253) (85,1.02319) (86,0.99307) (87,1.08697)};
\addlegendentry{Qwen3-32B (v3)}
\addplot[aubnavy, thick, mark=none] coordinates {(1,2.0705) (2,1.81139) (3,2.01331) (4,1.77979) (5,1.66656) (6,1.71452) (7,1.70885) (8,1.61271) (9,1.60254) (10,1.64993) (11,1.56095) (12,1.28249) (13,1.48842) (14,1.51418) (15,1.52652) (16,1.62052) (17,1.31821) (18,1.50759) (19,1.40676) (20,1.4294) (21,1.27951) (22,1.43485) (23,1.3776) (24,1.40614) (25,1.21036) (26,1.44231) (27,1.42276) (28,1.25789) (29,1.2714) (30,1.36176) (31,1.31025) (32,1.33326) (33,1.3029) (34,1.41087) (35,1.40428) (36,1.22433) (37,1.29868) (38,1.2698) (39,1.22121) (40,1.35826) (41,1.24048) (42,1.31156) (43,1.32634) (44,1.21078) (45,1.32593) (46,1.22074) (47,1.24678) (48,1.11509) (49,1.1523) (50,1.14668) (51,1.17207) (52,1.10703) (53,1.31714) (54,1.15476) (55,1.23495) (56,1.23628) (57,1.15578) (58,1.15414) (59,1.16291) (60,1.15116) (61,1.16002) (62,1.18633) (63,1.1712) (64,1.22961) (65,1.17977) (66,1.17975) (67,1.16636) (68,1.19368) (69,1.1595) (70,1.22438) (71,1.22661) (72,1.08794) (73,1.19145) (74,1.09367) (75,1.25394) (76,1.24677) (77,1.22429) (78,1.18221) (79,1.11779) (80,1.0903) (81,1.175) (82,1.11062) (83,1.10842) (84,1.14331) (85,1.21665) (86,1.18321) (87,1.2932)};
\addlegendentry{Qwen3-8B}
\addplot[aubgrey, thick, mark=none] coordinates {(1,2.61457) (2,2.19832) (3,2.36331) (4,2.10219) (5,1.99637) (6,1.97292) (7,1.95429) (8,1.86873) (9,1.87386) (10,1.91428) (11,1.79774) (12,1.52231) (13,1.75362) (14,1.77997) (15,1.8066) (16,1.87805) (17,1.58253) (18,1.76204) (19,1.67385) (20,1.67197) (21,1.52219) (22,1.68385) (23,1.61805) (24,1.65232) (25,1.47141) (26,1.73615) (27,1.66918) (28,1.50083) (29,1.54187) (30,1.61054) (31,1.58177) (32,1.5963) (33,1.56217) (34,1.68556) (35,1.64622) (36,1.49984) (37,1.59053) (38,1.52319) (39,1.48057) (40,1.62923) (41,1.51354) (42,1.58604) (43,1.60889) (44,1.49164) (45,1.60092) (46,1.48569) (47,1.50952) (48,1.38224) (49,1.41291) (50,1.40534) (51,1.42357) (52,1.36503) (53,1.60225) (54,1.43766) (55,1.48768) (56,1.51977) (57,1.4217) (58,1.44531) (59,1.4444) (60,1.42177) (61,1.42578) (62,1.44678) (63,1.44842) (64,1.52144) (65,1.4749) (66,1.43315) (67,1.43451) (68,1.48433) (69,1.40965) (70,1.48089) (71,1.50014) (72,1.34365) (73,1.47119) (74,1.35431) (75,1.51268) (76,1.57595) (77,1.51506) (78,1.47309) (79,1.38878) (80,1.35697) (81,1.45244) (82,1.38066) (83,1.39332) (84,1.43138) (85,1.48461) (86,1.45237) (87,1.59443)};
\addlegendentry{Qwen3-1.7B}
\nextgroupplot[title={(b) Held-out loss (2 whole runs, leakage-safe)}, ylabel={cross-entropy}, ymin=0.9, ymax=2.4]
\addplot[auburl, thick, mark=*, mark size=1.1] coordinates {(10,1.42625) (20,1.39031) (30,1.36437) (40,1.35773) (50,1.33574) (60,1.3253) (70,1.32298) (80,1.32261) (87,1.32211)};
\addlegendentry{Llama-3-8B-Instruct}
\addplot[auborange, thick, mark=*, mark size=1.1] coordinates {(10,1.23313) (20,1.18855) (30,1.15291) (40,1.13705) (50,1.11436) (60,1.10356) (70,1.10074) (80,1.09986) (87,1.09926)};
\addlegendentry{Gemma-3-27B-it}
\addplot[aubgreen, thick, mark=*, mark size=1.1] coordinates {(10,1.2284) (20,1.17547) (30,1.15062) (40,1.15051) (50,1.14199) (60,1.13439) (70,1.13362) (80,1.13319) (87,1.13317)};
\addlegendentry{Qwen3-32B (v3)}
\addplot[aubnavy, thick, mark=*, mark size=1.1] coordinates {(10,1.45875) (20,1.37452) (30,1.34361) (40,1.33461) (50,1.32496) (60,1.31964) (70,1.31735) (80,1.3179) (87,1.31765)};
\addlegendentry{Qwen3-8B}
\addplot[aubgrey, thick, mark=*, mark size=1.1] coordinates {(10,1.81486) (20,1.71984) (30,1.69829) (40,1.69085) (50,1.68308) (60,1.67591) (70,1.67531) (80,1.67481) (87,1.67512)};
\addlegendentry{Qwen3-1.7B}
\end{groupplot}\end{tikzpicture}
\caption{\textbf{LoRA-SFT optimisation on five backbones}, identical data, recipe and schedule (\cref{tab:hyper}); one adapter per backbone. Loss is masked to assistant tokens, so it measures only the target $\langle\texttt{think}\rangle$+JSON. Held-out loss is evaluated on two \emph{whole} archived runs withheld from training, so no document appears on both sides. Llama-3-8B: $1.81\!\to\!1.20$ train, best held-out $1.322$; Gemma-3-27B: $2.49\!\to\!0.86$ train, best held-out $1.099$; Qwen3-32B (v3): $1.57\!\to\!1.09$ train, best held-out $1.133$; Qwen3-8B: $2.07\!\to\!1.29$ train, best held-out $1.317$; Qwen3-1.7B: $2.61\!\to\!1.59$ train, best held-out $1.675$. Every curve is read from the archived per-adapter training logs.}\label{fig:losscurves}\end{figure}
}

\newcommand{\figRecipe}{%

\begin{figure}[H]\centering
\begin{tikzpicture}
\begin{axis}[width=0.96\columnwidth, height=5.0cm,
  xlabel={optimiser step}, ylabel={cross-entropy},
  tick label style={font=\scriptsize}, label style={font=\footnotesize},
  grid=both, grid style={aubline,very thin}, axis line style={aubgrey},
  legend style={font=\scriptsize, draw=aubline, fill=white, at={(0.98,0.98)},
                anchor=north east, row sep=-1pt}, legend columns=2,
  legend cell align=left, ymin=0.75, ymax=1.75]
\addplot[auburl, thick, mark=none] coordinates {(1,1.26199) (2,1.30018) (3,1.17595) (4,1.3337) (5,1.20353) (6,1.12223) (7,1.12994) (8,1.11119) (9,1.23221) (10,1.05872) (11,1.15585) (12,1.11195) (13,1.10122) (14,1.20387) (15,1.03517) (16,1.09329) (17,1.06384) (18,1.14923) (19,1.11114) (20,0.98446) (21,1.06968) (22,1.16604) (23,1.08564) (24,1.04948) (25,1.12111) (26,0.987555) (27,1.07782)};
\addlegendentry{v1 train}
\addplot[auburl, densely dashed, thick, mark=square*, mark size=1] coordinates {(3,1.30396) (6,1.26123) (9,1.22014) (12,1.19603) (15,1.18093) (18,1.17115) (21,1.16602) (24,1.16311) (27,1.16312)};
\addlegendentry{v1 held-out}
\addplot[auborange, thick, mark=none] coordinates {(1,1.26975) (2,1.23215) (3,1.33459) (4,1.14732) (5,1.23463) (6,1.11336) (7,1.17156) (8,1.11373) (9,1.11941) (10,1.10016) (11,1.06215) (12,1.1228) (13,1.30781) (14,1.04278) (15,1.05924) (16,1.07246) (17,1.06647) (18,1.05012) (19,1.09721) (20,1.01198) (21,0.98082) (22,1.15651) (23,1.02327) (24,0.996803) (25,0.960188) (26,0.972381) (27,1.11866) (28,0.974727) (29,0.94235) (30,0.97447) (31,0.994777) (32,0.930059) (33,1.00331) (34,0.997641) (35,0.967851) (36,1.01294) (37,0.982786) (38,1.09621) (39,0.886065)};
\addlegendentry{v2 train}
\addplot[auborange, densely dashed, thick, mark=square*, mark size=1] coordinates {(4,1.29542) (8,1.23157) (12,1.19251) (16,1.17107) (20,1.15591) (24,1.14516) (28,1.13821) (32,1.13461) (36,1.13454) (39,1.13404)};
\addlegendentry{v2 held-out}
\addplot[aubgreen, thick, mark=none] coordinates {(1,1.5683) (2,1.41641) (3,1.63971) (4,1.44831) (5,1.36435) (6,1.45998) (7,1.46829) (8,1.40202) (9,1.39051) (10,1.43629) (11,1.35228) (12,1.11062) (13,1.29119) (14,1.30316) (15,1.31612) (16,1.38224) (17,1.14043) (18,1.2974) (19,1.21606) (20,1.21579) (21,1.08362) (22,1.21129) (23,1.1614) (24,1.18317) (25,1.01846) (26,1.21671) (27,1.18247) (28,1.05057) (29,1.04936) (30,1.12307) (31,1.08008) (32,1.0973) (33,1.06817) (34,1.16737) (35,1.14687) (36,1.0155) (37,1.08526) (38,1.05244) (39,0.996595) (40,1.1321) (41,1.02833) (42,1.08512) (43,1.10436) (44,1.02689) (45,1.10755) (46,1.01521) (47,1.05033) (48,0.924842) (49,0.967136) (50,0.958768) (51,0.98736) (52,0.923124) (53,1.11251) (54,0.96042) (55,1.02808) (56,1.04149) (57,0.967938) (58,0.988062) (59,0.967542) (60,0.96656) (61,0.97696) (62,0.983795) (63,0.971302) (64,1.03425) (65,0.990996) (66,0.981328) (67,0.963156) (68,1.00149) (69,0.968312) (70,1.02737) (71,1.03095) (72,0.897424) (73,1.00173) (74,0.919326) (75,1.05063) (76,1.06) (77,1.03681) (78,0.988547) (79,0.936945) (80,0.907277) (81,0.979404) (82,0.943977) (83,0.926088) (84,0.959253) (85,1.02319) (86,0.99307) (87,1.08697)};
\addlegendentry{v3 train}
\addplot[aubgreen, densely dashed, thick, mark=square*, mark size=1] coordinates {(10,1.2284) (20,1.17547) (30,1.15062) (40,1.15051) (50,1.14199) (60,1.13439) (70,1.13362) (80,1.13319) (87,1.13317)};
\addlegendentry{v3 held-out}
\end{axis}\end{tikzpicture}
\caption{\textbf{Three SFT recipes on one fixed backbone} (Qwen3-32B), showing that the gain came from data volume, not from a longer run. v1 and v2 take 27 and 39 steps; v3 takes 87. Best held-out loss falls 1.163 \ensuremath{\to} 1.134 \ensuremath{\to} 1.133. v2 reaches the lowest \emph{training} loss (0.886) yet not the lowest held-out loss --- the classic small-corpus overfitting signature, and the reason early stopping (patience 3) reloads the best checkpoint rather than the last.}\label{fig:recipe}\end{figure}
}

\newcommand{\figDPO}{%

\begin{figure}[H]\centering
\begin{tikzpicture}
\begin{groupplot}[
  group style={group size=1 by 3, vertical sep=1.1cm,
               xlabels at=edge bottom, xticklabels at=edge bottom},
  width=0.92\columnwidth, height=3.5cm, xlabel={optimiser step},
  tick label style={font=\scriptsize}, label style={font=\footnotesize},
  title style={font=\small\bfseries}, grid=both,
  grid style={aubline,very thin}, axis line style={aubgrey},
  legend style={font=\tiny, draw=aubline, fill=white, row sep=-2pt},
  legend cell align=left,
]
\nextgroupplot[title={(a) DPO loss}, ylabel={$-\log\sigma(\beta\Delta)$}, ymin=0, ymax=0.75, legend style={at={(0.98,0.98)},anchor=north east}]
\addplot[auburl, thick, mark=none] coordinates {(1,0.693147) (2,0.693147) (3,0.660788) (4,0.608828) (5,0.596015) (6,0.524921) (7,0.498049) (8,0.525657) (9,0.392457) (10,0.447074) (11,0.36573) (12,0.388834) (13,0.264008) (14,0.335164) (15,0.267478) (16,0.28316) (17,0.230242) (18,0.26023) (19,0.255481) (20,0.266339) (21,0.204342) (22,0.203212) (23,0.135919) (24,0.187911) (25,0.193001) (26,0.178398) (27,0.0941701) (28,0.134509) (29,0.0883028) (30,0.145581) (31,0.019289) (32,0.0743045) (33,0.103937) (34,0.0653942) (35,0.0969262) (36,0.140477) (37,0.123111) (38,0.0908784) (39,0.105717) (40,0.0917199) (41,0.0640135) (42,0.0794723) (43,0.0665301) (44,0.029463) (45,0.0956946) (46,0.0755296) (47,0.0378381) (48,0.12413) (49,0.0496318) (50,0.0622363) (51,0.0562991) (52,0.0405114) (53,0.0736619) (54,0.12395) (55,0.0455078) (56,0.0687011) (57,0.0422023) (58,0.0685302) (59,0.0545611) (60,0.0815005) (61,0.0376791) (62,0.00458586) (63,0.0430582) (64,0.0646862) (65,0.0486114) (66,0.0487379) (67,0.0794556) (68,0.0566058) (69,0.0512194) (70,0.116302) (71,0.043974) (72,0.0639028) (73,0.0359994) (74,0.0349359) (75,0.0408698) (76,0.0379315) (77,0.0655017) (78,0.0416471) (79,0.0411443) (80,0.0388186) (81,0.0444121) (82,0.0458238) (83,0.0315821) (84,0.041552) (85,0.13063) (86,0.0604125) (87,0.0507527) (88,0.1414) (89,0.0401441) (90,0.0261602) (91,0.0344678) (92,0.0422945) (93,0.0809306)};
\addlegendentry{train}
\addplot[auborange, thick, densely dashed, mark=*, mark size=1] coordinates {(11,0.367238) (22,0.197586) (33,0.132559) (44,0.106256) (55,0.105326) (66,0.0954563) (77,0.0894784) (88,0.0916317) (93,0.0964708)};
\addlegendentry{held-out}
\draw[aubgrey,dashed] (axis cs:0,0.6931) -- (axis cs:93,0.6931);
\node[font=\tiny,aubgrey,anchor=north west] at (axis cs:3,0.675) {$\ln 2$: chance};
\nextgroupplot[title={(b) Implicit reward margin}, ylabel={$r_c-r_r$}, legend style={at={(0.02,0.98)},anchor=north west}]
\addplot[aubgreen, thick, mark=none] coordinates {(1,0) (2,0) (3,0.0677643) (4,0.18674) (5,0.213583) (6,0.374997) (7,0.459222) (8,0.374839) (9,0.78965) (10,0.623975) (11,0.857318) (12,0.804364) (13,1.31675) (14,0.942341) (15,1.34406) (16,1.19185) (17,1.47333) (18,1.41028) (19,1.26875) (20,1.44265) (21,1.67061) (22,1.86019) (23,2.0115) (24,1.69273) (25,1.68713) (26,1.72099) (27,2.44019) (28,2.18212) (29,2.60312) (30,2.19608) (31,3.93856) (32,2.93464) (33,2.36567) (34,3.12566) (35,2.78585) (36,2.50311) (37,2.43385) (38,2.66019) (39,2.2923) (40,2.64912) (41,2.88724) (42,2.53663) (43,3.48085) (44,3.78784) (45,2.59347) (46,2.88602) (47,3.48118) (48,2.98845) (49,3.04371) (50,2.98874) (51,3.11617) (52,3.37904) (53,2.76599) (54,2.96166) (55,3.38233) (56,2.79341) (57,3.67936) (58,3.30471) (59,3.0667) (60,2.6618) (61,3.51012) (62,5.38248) (63,3.49369) (64,2.82231) (65,3.56633) (66,3.30056) (67,2.9326) (68,3.11442) (69,3.06672) (70,3.18974) (71,3.34649) (72,3.25987) (73,3.72963) (74,3.65908) (75,3.83193) (76,3.33939) (77,3.20155) (78,3.35344) (79,3.72993) (80,3.72165) (81,3.40616) (82,3.88543) (83,3.84191) (84,3.41022) (85,3.25325) (86,2.93667) (87,3.33778) (88,2.87954) (89,3.43146) (90,3.7756) (91,3.56438) (92,3.77045) (93,2.47343)};
\addlegendentry{margin}
\addplot[auburl, thin, mark=none] coordinates {(1,0) (2,0) (3,0.0197464) (4,0.0688553) (5,-0.0763062) (6,0.0744858) (7,-0.0263138) (8,-0.0774506) (9,0.0716736) (10,-0.0541771) (11,0.0729095) (12,-0.0823227) (13,-0.132672) (14,-0.0304871) (15,0.0185997) (16,0.0283531) (17,-0.014209) (18,-0.094709) (19,-0.261005) (20,-0.105928) (21,-0.0168381) (22,0.0372696) (23,-0.0892654) (24,-0.398615) (25,-0.0861969) (26,-0.144763) (27,-0.256332) (28,-0.305353) (29,-0.338225) (30,-0.324358) (31,-0.16449) (32,-0.201504) (33,-0.196581) (34,-0.430908) (35,-0.228831) (36,-0.199487) (37,-0.133269) (38,-0.522752) (39,-0.281702) (40,-0.761867) (41,-0.51497) (42,-0.374567) (43,-0.245399) (44,-0.618941) (45,-0.361173) (46,-0.623581) (47,-0.338529) (48,-0.479494) (49,-0.492878) (50,-0.748895) (51,-0.389476) (52,-0.907218) (53,-0.851596) (54,-0.67589) (55,-0.771365) (56,-0.662673) (57,-0.766725) (58,-0.460839) (59,-0.67448) (60,-0.829799) (61,-1.04854) (62,-0.180737) (63,-0.576975) (64,-0.473055) (65,-0.770341) (66,-0.67102) (67,-0.488538) (68,-0.630091) (69,-0.767027) (70,-0.534813) (71,-0.448164) (72,-0.531624) (73,-0.7353) (74,-0.709862) (75,-0.70186) (76,-0.482166) (77,-0.700939) (78,-1.07155) (79,-0.59116) (80,-0.842033) (81,-0.826326) (82,-0.612976) (83,-0.837334) (84,-0.788226) (85,-0.751917) (86,-1.02888) (87,-1.08118) (88,-0.564981) (89,-0.768031) (90,-1.01691) (91,-0.787624) (92,-0.682718) (93,-1.97906)};
\addlegendentry{$r_{\text{chosen}}$}
\addplot[auborange, thin, mark=none] coordinates {(1,0) (2,0) (3,-0.0480179) (4,-0.117885) (5,-0.289889) (6,-0.300511) (7,-0.485536) (8,-0.45229) (9,-0.717976) (10,-0.678152) (11,-0.784409) (12,-0.886687) (13,-1.44942) (14,-0.972828) (15,-1.32546) (16,-1.16349) (17,-1.48754) (18,-1.50499) (19,-1.52976) (20,-1.54857) (21,-1.68745) (22,-1.82292) (23,-2.10076) (24,-2.09134) (25,-1.77333) (26,-1.86575) (27,-2.69652) (28,-2.48747) (29,-2.94135) (30,-2.52044) (31,-4.10305) (32,-3.13614) (33,-2.56225) (34,-3.55657) (35,-3.01469) (36,-2.7026) (37,-2.56712) (38,-3.18294) (39,-2.574) (40,-3.41099) (41,-3.40221) (42,-2.91119) (43,-3.72625) (44,-4.40678) (45,-2.95464) (46,-3.5096) (47,-3.81971) (48,-3.46794) (49,-3.53659) (50,-3.73763) (51,-3.50565) (52,-4.28626) (53,-3.61758) (54,-3.63755) (55,-4.1537) (56,-3.45609) (57,-4.44608) (58,-3.76555) (59,-3.74118) (60,-3.4916) (61,-4.55866) (62,-5.56322) (63,-4.07067) (64,-3.29537) (65,-4.33667) (66,-3.97158) (67,-3.42114) (68,-3.74452) (69,-3.83375) (70,-3.72456) (71,-3.79465) (72,-3.7915) (73,-4.46493) (74,-4.36894) (75,-4.53379) (76,-3.82155) (77,-3.90249) (78,-4.42499) (79,-4.32109) (80,-4.56368) (81,-4.23249) (82,-4.4984) (83,-4.67924) (84,-4.19845) (85,-4.00516) (86,-3.96555) (87,-4.41896) (88,-3.44452) (89,-4.19949) (90,-4.79252) (91,-4.352) (92,-4.45317) (93,-4.45249)};
\addlegendentry{$r_{\text{rejected}}$}
\nextgroupplot[title={(c) Preference accuracy}, ylabel={fraction correct}, ymin=0, ymax=1.05, legend style={at={(0.98,0.02)},anchor=south east}]
\addplot[aubnavy, thick, mark=none] coordinates {(1,0) (2,0) (3,0.75) (4,0.75) (5,0.875) (6,1) (7,1) (8,1) (9,1) (10,1) (11,1) (12,1) (13,1) (14,1) (15,1) (16,1) (17,1) (18,1) (19,1) (20,0.875) (21,1) (22,0.875) (23,1) (24,1) (25,1) (26,1) (27,1) (28,1) (29,1) (30,1) (31,1) (32,1) (33,1) (34,1) (35,1) (36,0.875) (37,1) (38,1) (39,1) (40,1) (41,1) (42,1) (43,1) (44,1) (45,1) (46,1) (47,1) (48,0.875) (49,1) (50,1) (51,1) (52,1) (53,1) (54,0.875) (55,1) (56,1) (57,1) (58,1) (59,1) (60,1) (61,1) (62,1) (63,1) (64,1) (65,1) (66,1) (67,1) (68,1) (69,1) (70,0.875) (71,1) (72,1) (73,1) (74,1) (75,1) (76,1) (77,1) (78,1) (79,1) (80,1) (81,1) (82,1) (83,1) (84,1) (85,0.875) (86,1) (87,1) (88,0.875) (89,1) (90,1) (91,1) (92,1) (93,1)};
\addlegendentry{$\Pr[r_c>r_r]$}
\draw[aubgrey,dashed] (axis cs:0,0.5) -- (axis cs:93,0.5);
\end{groupplot}\end{tikzpicture}
\caption{\textbf{DPO calibration pass} on top of the SFT adapter ($\beta=0.1$), the run that targets the over-doom bias. Loss falls from the $\ln 2\approx0.693$ chance level to 0.081 over 93 steps (3641\,s), best held-out 0.0895. Panel (b) separates the implicit rewards $r=\beta\log\frac{\pi_\theta}{\pi_{\text{ref}}}$: the margin widens because the rejected branch is pushed down, the diagnostic signature of a preference pass that is learning the contrast rather than merely raising likelihood everywhere. Panel (c) is the fraction of pairs ranked correctly within each \emph{training} batch of eight, which is why it moves in steps of $1/8$; on the $26$ held-out pairs the same quantity is $26/26$ at six of nine evaluations and never below $25/26$. \emph{These dynamics are healthy but the downstream effect is null} (\cref{tab:gold50}): DPO leaves Gold-50 accuracy unchanged, which we read as the preference set (over-doom, invalid-JSON, hallucination triples) being too small and too synthetic to move judgment.}\label{fig:dpo}\end{figure}
}

\newcommand{\figLeaderboard}{%

\begin{figure}[tbp]\centering
\begin{tikzpicture}
\begin{axis}[width=0.66\columnwidth, height=7.4cm,
  xlabel={accuracy vs.\ verified outcome (\%)}, xmin=0, xmax=112,
  xtick={0,20,40,60,80,100},
  y=0.46cm, ymin=-0.7, ymax=10.7,
  ytick={0,1,2,3,4,5,6,7,8,9,10},
  yticklabels={
{Qwen3-1.7B + LoRA-SFT},{Llama-3-8B-Instruct},{Gemma-3-27B-it},{Qwen3-1.7B},{Qwen3-8B + LoRA-SFT},{Gemma-3-27B + LoRA-SFT},{Qwen3-8B},{Qwen3-32B},{Llama-3-8B + LoRA-SFT},{Qwen3-32B + SFT\ensuremath{\to}DPO},{Qwen3-32B + LoRA-SFT}
},
  yticklabel style={font=\tiny}, tick label style={font=\scriptsize},
  label style={font=\footnotesize}, xmajorgrids,
  grid style={aubline,very thin}, axis line style={aubgrey},
  clip mode=individual,
]
\draw[aubgrey,dashed] (axis cs:20,-0.6) -- (axis cs:20,10.6);
\node[font=\tiny,aubgrey,anchor=south] at (axis cs:20,10.45) {chance $=20\%$};
\addplot[xbar, fill=auburl!70, draw=auburl, bar width=0.26cm, error bars/x dir=both, error bars/x explicit] coordinates {(6,0) +- (3.9,10.2)};
\node[font=\tiny,aubnavy,anchor=west] at (axis cs:16.2,0) {\,6};
\addplot[xbar, fill=aubgrey!35, draw=aubgrey, bar width=0.26cm, error bars/x dir=both, error bars/x explicit] coordinates {(16,1) +- (7.7,12.5)};
\node[font=\tiny,aubnavy,anchor=west] at (axis cs:28.5,1) {\,16};
\addplot[xbar, fill=aubgrey!35, draw=aubgrey, bar width=0.26cm, error bars/x dir=both, error bars/x explicit] coordinates {(36,2) +- (11.9,13.9)};
\node[font=\tiny,aubnavy,anchor=west] at (axis cs:49.9,2) {\,36};
\addplot[xbar, fill=aubgrey!35, draw=aubgrey, bar width=0.26cm, error bars/x dir=both, error bars/x explicit] coordinates {(46,3) +- (13,13.6)};
\node[font=\tiny,aubnavy,anchor=west] at (axis cs:59.6,3) {\,46};
\addplot[xbar, fill=auburl!70, draw=auburl, bar width=0.26cm, error bars/x dir=both, error bars/x explicit] coordinates {(50,4) +- (13.4,13.4)};
\node[font=\tiny,aubnavy,anchor=west] at (axis cs:63.4,4) {\,50};
\addplot[xbar, fill=auburl!70, draw=auburl, bar width=0.26cm, error bars/x dir=both, error bars/x explicit] coordinates {(50,5) +- (13.4,13.4)};
\node[font=\tiny,aubnavy,anchor=west] at (axis cs:63.4,5) {\,50};
\addplot[xbar, fill=aubgrey!35, draw=aubgrey, bar width=0.26cm, error bars/x dir=both, error bars/x explicit] coordinates {(54,6) +- (13.6,13)};
\node[font=\tiny,aubnavy,anchor=west] at (axis cs:67,6) {\,54};
\addplot[xbar, fill=aubgrey!35, draw=aubgrey, bar width=0.26cm, error bars/x dir=both, error bars/x explicit] coordinates {(56,7) +- (13.7,12.8)};
\node[font=\tiny,aubnavy,anchor=west] at (axis cs:68.8,7) {\,56};
\addplot[xbar, fill=auburl!70, draw=auburl, bar width=0.26cm, error bars/x dir=both, error bars/x explicit] coordinates {(60,8) +- (13.8,12.4)};
\node[font=\tiny,aubnavy,anchor=west] at (axis cs:72.4,8) {\,60};
\addplot[xbar, fill=aubgreen!70, draw=aubgreen, bar width=0.26cm, error bars/x dir=both, error bars/x explicit] coordinates {(62,9) +- (13.8,12.1)};
\node[font=\tiny,aubnavy,anchor=west] at (axis cs:74.1,9) {\,62};
\addplot[xbar, fill=auburl!70, draw=auburl, bar width=0.26cm, error bars/x dir=both, error bars/x explicit] coordinates {(62,10) +- (13.8,12.1)};
\node[font=\tiny,aubnavy,anchor=west] at (axis cs:74.1,10) {\,62};
\end{axis}\end{tikzpicture}
\caption{\textbf{Gold-50 leaderboard.} Accuracy of the five-way release verdict against the \emph{verified} real-world outcome on $n=50$ cases, balanced at 10 per class, all 50/50 labels manually re-verified (\cref{sec:goldset}). Bars are teal for LoRA-SFT adapters, green for SFT\ensuremath{\to}DPO and grey for untuned bases; whiskers are 95\% Wilson score intervals, which are wide at this $n$ and overlap for most pairs --- hence the paired McNemar tests in \cref{tab:paired}, which condition on the same cases. Unparseable outputs are scored as wrong, the conservative choice; \cref{tab:gold50} also reports the accuracy conditional on in-schema output.}\label{fig:leaderboard}\end{figure}
}

\newcommand{\figPaired}{%

\begin{figure}[tbp]\centering
\begin{tikzpicture}
\begin{axis}[width=0.99\columnwidth, height=5.0cm,
  ybar, bar width=0.26cm, ymin=0, ymax=97,
  ytick={0,20,40,60,80},
  ylabel={Gold-50 accuracy (\%)}, symbolic x coords={
Qwen3-1.7B,Llama-3-8B,Qwen3-8B,Gemma-3-27B,Qwen3-32B
},
  xtick=data, xticklabels={{Qwen3-1.7B},{Llama-3-8B-Instruct},{Qwen3-8B},{Gemma-3-27B-it},{Qwen3-32B}}, xticklabel style={font=\scriptsize, rotate=18, anchor=north east},
  tick label style={font=\scriptsize}, label style={font=\footnotesize},
  ymajorgrids, grid style={aubline,very thin}, axis line style={aubgrey},
  legend style={font=\scriptsize, draw=aubline, fill=white, at={(0.5,1.04)},
                anchor=south, legend columns=2,
                /tikz/every even column/.append style={column sep=1.2ex}},
  legend cell align=left,
  nodes near coords={\pgfmathprintnumber[fixed,precision=0]{\pgfplotspointmeta}},
  every node near coord/.append style={font=\tiny, anchor=south},
  clip=false, enlarge x limits=0.13,
]
\draw[aubgrey,dashed] (rel axis cs:0,0.2062) -- (rel axis cs:1,0.2062);
\addplot[fill=aubgrey!35, draw=aubgrey, point meta=explicit] coordinates {(Qwen3-1.7B,46) [46] (Llama-3-8B,16) [16] (Qwen3-8B,54) [54] (Gemma-3-27B,36) [36] (Qwen3-32B,56) [56]};
\addlegendentry{untuned base}
\addplot[fill=auburl!70, draw=auburl, point meta=explicit] coordinates {(Qwen3-1.7B,6) [6] (Llama-3-8B,60) [60] (Qwen3-8B,50) [50] (Gemma-3-27B,50) [50] (Qwen3-32B,62) [62]};
\addlegendentry{+ LoRA-SFT}
\node[font=\tiny,auborange] at (axis cs:Qwen3-1.7B,91) {$p{<}0.001$};
\node[font=\tiny,auborange] at (axis cs:Qwen3-1.7B,83) {-40\,pp};
\node[font=\tiny,auborange] at (axis cs:Llama-3-8B,91) {$p{<}0.001$};
\node[font=\tiny,auborange] at (axis cs:Llama-3-8B,83) {+44\,pp};
\node[font=\tiny,aubgrey] at (axis cs:Qwen3-8B,91) {$p{=}0.81$};
\node[font=\tiny,aubgrey] at (axis cs:Qwen3-8B,83) {-4\,pp};
\node[font=\tiny,aubgrey] at (axis cs:Gemma-3-27B,91) {$p{=}0.21$};
\node[font=\tiny,aubgrey] at (axis cs:Gemma-3-27B,83) {+14\,pp};
\node[font=\tiny,aubgrey] at (axis cs:Qwen3-32B,91) {$p{=}0.61$};
\node[font=\tiny,aubgrey] at (axis cs:Qwen3-32B,83) {+6\,pp};
\end{axis}\end{tikzpicture}
\caption{\textbf{The same fine-tune helps or destroys, depending on the backbone.} Each pair is one backbone evaluated before and after LoRA-SFT on \emph{identical} data, recipe, seed and prompts, scored on the \emph{same} 50 cases; $p$ is a two-sided exact McNemar test on the discordant pairs (\cref{tab:paired}), which is the correct test here because the two systems see the same items. The effect ranges from $+44$\,pp (Llama-3-8B) to $-40$\,pp (Qwen3-1.7B), and only those two extremes are significant --- the three mid-size gains are not distinguishable from noise at $n=50$. Dashed line is the $20\%$ five-way chance level.}\label{fig:paired}\end{figure}
}

\newcommand{\figConfusion}{%

\begin{figure*}[tp]\centering
\begin{tikzpicture}
\begin{groupplot}[
  group style={group size=3 by 1, horizontal sep=1.35cm,
               ylabels at=edge left, yticklabels at=edge left},
  width=0.62\columnwidth, height=4.2cm,
  colormap={aub}{color=(white) color=(aubteal) color=(auburl)},
  xlabel={predicted}, ylabel={verified outcome},
  tick label style={font=\scriptsize}, label style={font=\footnotesize},
  title style={font=\small\bfseries},
  xticklabel style={rotate=35, anchor=north east},
  enlarge x limits=false, enlarge y limits=false, y dir=reverse,
  x=0.6cm, y=0.6cm,
]
\nextgroupplot[title={(a) Qwen3-32B base},
  xtick={0,1,2,3,4,5},
  xticklabels={ship,revise,delay,segment,mitigate,none},
  ytick={0,1,2,3,4},
  yticklabels={ship,revise,delay,segment,mitigate},
  point meta min=0, point meta max=10]
\addplot[matrix plot*, mesh/cols=6, point meta=explicit] table[meta=C, row sep=crcr] {x y C\\0 0 3\\1 0 1\\2 0 0\\3 0 2\\4 0 4\\5 0 0\\0 1 1\\1 1 6\\2 1 0\\3 1 1\\4 1 2\\5 1 0\\0 2 0\\1 2 2\\2 2 8\\3 2 0\\4 2 0\\5 2 0\\0 3 3\\1 3 0\\2 3 0\\3 3 4\\4 3 3\\5 3 0\\0 4 2\\1 4 1\\2 4 0\\3 4 0\\4 4 7\\5 4 0\\};
\node[font=\tiny,aubnavy] at (axis cs:0,0) {3};
\node[font=\tiny,aubnavy] at (axis cs:1,0) {1};
\node[font=\tiny,aubnavy] at (axis cs:3,0) {2};
\node[font=\tiny,aubnavy] at (axis cs:4,0) {4};
\node[font=\tiny,aubnavy] at (axis cs:0,1) {1};
\node[font=\tiny,white] at (axis cs:1,1) {6};
\node[font=\tiny,aubnavy] at (axis cs:3,1) {1};
\node[font=\tiny,aubnavy] at (axis cs:4,1) {2};
\node[font=\tiny,aubnavy] at (axis cs:1,2) {2};
\node[font=\tiny,white] at (axis cs:2,2) {8};
\node[font=\tiny,aubnavy] at (axis cs:0,3) {3};
\node[font=\tiny,aubnavy] at (axis cs:3,3) {4};
\node[font=\tiny,aubnavy] at (axis cs:4,3) {3};
\node[font=\tiny,aubnavy] at (axis cs:0,4) {2};
\node[font=\tiny,aubnavy] at (axis cs:1,4) {1};
\node[font=\tiny,white] at (axis cs:4,4) {7};
\nextgroupplot[title={(b) Qwen3-32B + LoRA-SFT},
  xtick={0,1,2,3,4,5},
  xticklabels={ship,revise,delay,segment,mitigate,none},
  ytick={0,1,2,3,4},
  yticklabels={},
  point meta min=0, point meta max=10]
\addplot[matrix plot*, mesh/cols=6, point meta=explicit] table[meta=C, row sep=crcr] {x y C\\0 0 3\\1 0 4\\2 0 0\\3 0 0\\4 0 3\\5 0 0\\0 1 2\\1 1 8\\2 1 0\\3 1 0\\4 1 0\\5 1 0\\0 2 0\\1 2 1\\2 2 9\\3 2 0\\4 2 0\\5 2 0\\0 3 4\\1 3 2\\2 3 0\\3 3 2\\4 3 2\\5 3 0\\0 4 1\\1 4 0\\2 4 0\\3 4 0\\4 4 9\\5 4 0\\};
\node[font=\tiny,aubnavy] at (axis cs:0,0) {3};
\node[font=\tiny,aubnavy] at (axis cs:1,0) {4};
\node[font=\tiny,aubnavy] at (axis cs:4,0) {3};
\node[font=\tiny,aubnavy] at (axis cs:0,1) {2};
\node[font=\tiny,white] at (axis cs:1,1) {8};
\node[font=\tiny,aubnavy] at (axis cs:1,2) {1};
\node[font=\tiny,white] at (axis cs:2,2) {9};
\node[font=\tiny,aubnavy] at (axis cs:0,3) {4};
\node[font=\tiny,aubnavy] at (axis cs:1,3) {2};
\node[font=\tiny,aubnavy] at (axis cs:3,3) {2};
\node[font=\tiny,aubnavy] at (axis cs:4,3) {2};
\node[font=\tiny,aubnavy] at (axis cs:0,4) {1};
\node[font=\tiny,white] at (axis cs:4,4) {9};
\nextgroupplot[title={(c) Gemma-3-27B + LoRA-SFT},
  xtick={0,1,2,3,4,5},
  xticklabels={ship,revise,delay,segment,mitigate,none},
  ytick={0,1,2,3,4},
  yticklabels={},
  point meta min=0, point meta max=10]
\addplot[matrix plot*, mesh/cols=6, point meta=explicit] table[meta=C, row sep=crcr] {x y C\\0 0 1\\1 0 0\\2 0 0\\3 0 0\\4 0 9\\5 0 0\\0 1 0\\1 1 5\\2 1 0\\3 1 0\\4 1 5\\5 1 0\\0 2 0\\1 2 1\\2 2 9\\3 2 0\\4 2 0\\5 2 0\\0 3 2\\1 3 0\\2 3 0\\3 3 0\\4 3 8\\5 3 0\\0 4 0\\1 4 0\\2 4 0\\3 4 0\\4 4 10\\5 4 0\\};
\node[font=\tiny,aubnavy] at (axis cs:0,0) {1};
\node[font=\tiny,white] at (axis cs:4,0) {9};
\node[font=\tiny,aubnavy] at (axis cs:1,1) {5};
\node[font=\tiny,aubnavy] at (axis cs:4,1) {5};
\node[font=\tiny,aubnavy] at (axis cs:1,2) {1};
\node[font=\tiny,white] at (axis cs:2,2) {9};
\node[font=\tiny,aubnavy] at (axis cs:0,3) {2};
\node[font=\tiny,white] at (axis cs:4,3) {8};
\node[font=\tiny,white] at (axis cs:4,4) {10};
\end{groupplot}\end{tikzpicture}
\caption{\textbf{Confusion on the Gold-50 set}, rows are the verified outcome (10 each), columns the predicted action, \code{none} the unparseable bucket. The diagonal is accuracy; the informative structure is off-diagonal. All three systems concentrate mass in the \emph{mitigate} column, and the \emph{segment} column is nearly empty --- models rarely propose splitting a release by cohort even when that is what the organisation actually did. Fine-tuning (b vs a) sharpens the diagonal on \emph{ship} and \emph{revise} but does not create \emph{segment} behaviour, and on Gemma (c) it collapses almost entirely into \emph{mitigate}. \cref{fig:biasheat} quantifies this across all eleven systems.}\label{fig:confusion}\end{figure*}
}

\newcommand{\figHeatPair}{%

\begin{figure*}[tp]\centering
\begin{tikzpicture}
\begin{groupplot}[
  group style={group size=2 by 1, horizontal sep=0.9cm},
  width=0.9\columnwidth, height=6.6cm,
  tick label style={font=\scriptsize}, label style={font=\footnotesize},
  title style={font=\small\bfseries},
  xticklabel style={rotate=35, anchor=north east},
  enlarge x limits=false, enlarge y limits=false, y dir=reverse,
  ytick={0,1,2,3,4,5,6,7,8,9,10},
  yticklabels={{Qwen3-32B + LoRA-SFT},{Qwen3-32B + SFT\ensuremath{\to}DPO},{Llama-3-8B + LoRA-SFT},{Qwen3-32B},{Qwen3-8B},{Qwen3-8B + LoRA-SFT},{Gemma-3-27B + LoRA-SFT},{Qwen3-1.7B},{Gemma-3-27B-it},{Llama-3-8B-Instruct},{Qwen3-1.7B + LoRA-SFT}},
  y=0.42cm, x=0.72cm,
]
\nextgroupplot[title={(a) Per-class recall (out of 10)}, xlabel={verified outcome class}, colormap={aub}{color=(white) color=(aubteal) color=(auburl)},
  xtick={0,1,2,3,4},
  xticklabels={ship,revise,delay,segment,mitigate},
  point meta min=0, point meta max=10]
\addplot[matrix plot*, mesh/cols=5, point meta=explicit] table[meta=C, row sep=crcr] {x y C\\0 0 3\\1 0 8\\2 0 9\\3 0 2\\4 0 9\\0 1 3\\1 1 8\\2 1 9\\3 1 2\\4 1 9\\0 2 4\\1 2 8\\2 2 8\\3 2 1\\4 2 9\\0 3 3\\1 3 6\\2 3 8\\3 3 4\\4 3 7\\0 4 1\\1 4 5\\2 4 9\\3 4 3\\4 4 9\\0 5 5\\1 5 7\\2 5 10\\3 5 3\\4 5 0\\0 6 1\\1 6 5\\2 6 9\\3 6 0\\4 6 10\\0 7 8\\1 7 5\\2 7 10\\3 7 0\\4 7 0\\0 8 7\\1 8 4\\2 8 7\\3 8 0\\4 8 0\\0 9 4\\1 9 0\\2 9 4\\3 9 0\\4 9 0\\0 10 2\\1 10 0\\2 10 0\\3 10 1\\4 10 0\\};
\node[font=\tiny,aubnavy] at (axis cs:0,0) {3};
\node[font=\tiny,white] at (axis cs:1,0) {8};
\node[font=\tiny,white] at (axis cs:2,0) {9};
\node[font=\tiny,aubnavy] at (axis cs:3,0) {2};
\node[font=\tiny,white] at (axis cs:4,0) {9};
\node[font=\tiny,aubnavy] at (axis cs:0,1) {3};
\node[font=\tiny,white] at (axis cs:1,1) {8};
\node[font=\tiny,white] at (axis cs:2,1) {9};
\node[font=\tiny,aubnavy] at (axis cs:3,1) {2};
\node[font=\tiny,white] at (axis cs:4,1) {9};
\node[font=\tiny,aubnavy] at (axis cs:0,2) {4};
\node[font=\tiny,white] at (axis cs:1,2) {8};
\node[font=\tiny,white] at (axis cs:2,2) {8};
\node[font=\tiny,aubnavy] at (axis cs:3,2) {1};
\node[font=\tiny,white] at (axis cs:4,2) {9};
\node[font=\tiny,aubnavy] at (axis cs:0,3) {3};
\node[font=\tiny,white] at (axis cs:1,3) {6};
\node[font=\tiny,white] at (axis cs:2,3) {8};
\node[font=\tiny,aubnavy] at (axis cs:3,3) {4};
\node[font=\tiny,white] at (axis cs:4,3) {7};
\node[font=\tiny,aubnavy] at (axis cs:0,4) {1};
\node[font=\tiny,aubnavy] at (axis cs:1,4) {5};
\node[font=\tiny,white] at (axis cs:2,4) {9};
\node[font=\tiny,aubnavy] at (axis cs:3,4) {3};
\node[font=\tiny,white] at (axis cs:4,4) {9};
\node[font=\tiny,aubnavy] at (axis cs:0,5) {5};
\node[font=\tiny,white] at (axis cs:1,5) {7};
\node[font=\tiny,white] at (axis cs:2,5) {10};
\node[font=\tiny,aubnavy] at (axis cs:3,5) {3};
\node[font=\tiny,aubnavy] at (axis cs:0,6) {1};
\node[font=\tiny,aubnavy] at (axis cs:1,6) {5};
\node[font=\tiny,white] at (axis cs:2,6) {9};
\node[font=\tiny,white] at (axis cs:4,6) {10};
\node[font=\tiny,white] at (axis cs:0,7) {8};
\node[font=\tiny,aubnavy] at (axis cs:1,7) {5};
\node[font=\tiny,white] at (axis cs:2,7) {10};
\node[font=\tiny,white] at (axis cs:0,8) {7};
\node[font=\tiny,aubnavy] at (axis cs:1,8) {4};
\node[font=\tiny,white] at (axis cs:2,8) {7};
\node[font=\tiny,aubnavy] at (axis cs:0,9) {4};
\node[font=\tiny,aubnavy] at (axis cs:2,9) {4};
\node[font=\tiny,aubnavy] at (axis cs:0,10) {2};
\node[font=\tiny,aubnavy] at (axis cs:3,10) {1};
\nextgroupplot[title={(b) Predictions emitted (50 total)}, xlabel={predicted action}, yticklabels={}, colormap={aub2}{color=(white) color=(aubpeach) color=(auborange)},
  xtick={0,1,2,3,4,5},
  xticklabels={ship,revise,delay,segment,mitigate,none},
  point meta min=0, point meta max=35]
\addplot[matrix plot*, mesh/cols=6, point meta=explicit] table[meta=C, row sep=crcr] {x y C\\0 0 10\\1 0 15\\2 0 9\\3 0 2\\4 0 14\\5 0 0\\0 1 10\\1 1 15\\2 1 9\\3 1 2\\4 1 14\\5 1 0\\0 2 8\\1 2 9\\2 2 8\\3 2 1\\4 2 23\\5 2 1\\0 3 9\\1 3 10\\2 3 8\\3 3 7\\4 3 16\\5 3 0\\0 4 3\\1 4 7\\2 4 9\\3 4 5\\4 4 26\\5 4 0\\0 5 18\\1 5 9\\2 5 10\\3 5 3\\4 5 4\\5 5 6\\0 6 3\\1 6 6\\2 6 9\\3 6 0\\4 6 32\\5 6 0\\0 7 32\\1 7 6\\2 7 11\\3 7 0\\4 7 1\\5 7 0\\0 8 29\\1 8 6\\2 8 7\\3 8 0\\4 8 8\\5 8 0\\0 9 24\\1 9 0\\2 9 4\\3 9 0\\4 9 22\\5 9 0\\0 10 12\\1 10 0\\2 10 1\\3 10 2\\4 10 0\\5 10 35\\};
\node[font=\tiny,aubnavy] at (axis cs:0,0) {10};
\node[font=\tiny,aubnavy] at (axis cs:1,0) {15};
\node[font=\tiny,aubnavy] at (axis cs:2,0) {9};
\node[font=\tiny,aubnavy] at (axis cs:3,0) {2};
\node[font=\tiny,aubnavy] at (axis cs:4,0) {14};
\node[font=\tiny,aubnavy] at (axis cs:0,1) {10};
\node[font=\tiny,aubnavy] at (axis cs:1,1) {15};
\node[font=\tiny,aubnavy] at (axis cs:2,1) {9};
\node[font=\tiny,aubnavy] at (axis cs:3,1) {2};
\node[font=\tiny,aubnavy] at (axis cs:4,1) {14};
\node[font=\tiny,aubnavy] at (axis cs:0,2) {8};
\node[font=\tiny,aubnavy] at (axis cs:1,2) {9};
\node[font=\tiny,aubnavy] at (axis cs:2,2) {8};
\node[font=\tiny,aubnavy] at (axis cs:3,2) {1};
\node[font=\tiny,white] at (axis cs:4,2) {23};
\node[font=\tiny,aubnavy] at (axis cs:5,2) {1};
\node[font=\tiny,aubnavy] at (axis cs:0,3) {9};
\node[font=\tiny,aubnavy] at (axis cs:1,3) {10};
\node[font=\tiny,aubnavy] at (axis cs:2,3) {8};
\node[font=\tiny,aubnavy] at (axis cs:3,3) {7};
\node[font=\tiny,aubnavy] at (axis cs:4,3) {16};
\node[font=\tiny,aubnavy] at (axis cs:0,4) {3};
\node[font=\tiny,aubnavy] at (axis cs:1,4) {7};
\node[font=\tiny,aubnavy] at (axis cs:2,4) {9};
\node[font=\tiny,aubnavy] at (axis cs:3,4) {5};
\node[font=\tiny,white] at (axis cs:4,4) {26};
\node[font=\tiny,aubnavy] at (axis cs:0,5) {18};
\node[font=\tiny,aubnavy] at (axis cs:1,5) {9};
\node[font=\tiny,aubnavy] at (axis cs:2,5) {10};
\node[font=\tiny,aubnavy] at (axis cs:3,5) {3};
\node[font=\tiny,aubnavy] at (axis cs:4,5) {4};
\node[font=\tiny,aubnavy] at (axis cs:5,5) {6};
\node[font=\tiny,aubnavy] at (axis cs:0,6) {3};
\node[font=\tiny,aubnavy] at (axis cs:1,6) {6};
\node[font=\tiny,aubnavy] at (axis cs:2,6) {9};
\node[font=\tiny,white] at (axis cs:4,6) {32};
\node[font=\tiny,white] at (axis cs:0,7) {32};
\node[font=\tiny,aubnavy] at (axis cs:1,7) {6};
\node[font=\tiny,aubnavy] at (axis cs:2,7) {11};
\node[font=\tiny,aubnavy] at (axis cs:4,7) {1};
\node[font=\tiny,white] at (axis cs:0,8) {29};
\node[font=\tiny,aubnavy] at (axis cs:1,8) {6};
\node[font=\tiny,aubnavy] at (axis cs:2,8) {7};
\node[font=\tiny,aubnavy] at (axis cs:4,8) {8};
\node[font=\tiny,white] at (axis cs:0,9) {24};
\node[font=\tiny,aubnavy] at (axis cs:2,9) {4};
\node[font=\tiny,white] at (axis cs:4,9) {22};
\node[font=\tiny,aubnavy] at (axis cs:0,10) {12};
\node[font=\tiny,aubnavy] at (axis cs:2,10) {1};
\node[font=\tiny,aubnavy] at (axis cs:3,10) {2};
\node[font=\tiny,white] at (axis cs:5,10) {35};
\end{groupplot}\end{tikzpicture}
\caption{\textbf{What the systems get right, and what they say instead.} Rows are the eleven systems ordered by overall accuracy. (a) Recall per verified class, support 10 each: every system is strongest on the unambiguous ends (\emph{ship}, \emph{revise}) and weakest on \emph{segment}, where the best system recovers 4/10 and four systems recover none. (b) The marginal distribution of what each system \emph{emitted}, against a uniform truth of 10 per class --- a bias fingerprint that is independent of correctness. Values above 10 are over-prediction. Seven of the eleven systems over-predict \emph{mitigate}, by up to $3.2\times$ its true frequency (32/50, Gemma-3-27B+SFT), and it is the modal prediction of four of them --- the quantitative form of the ``over-doom'' failure this project has tracked since its local baseline, and the target of the DPO pass in \cref{fig:dpo}. The \emph{segment} column is the mirror image: eight systems emit that action at most twice in fifty attempts.}\label{fig:biasheat}\end{figure*}
}

\newcommand{\figDrivers}{%

\begin{figure}[H]\centering
\begin{tikzpicture}
\begin{axis}[width=0.97\columnwidth, height=5.2cm,
  ybar, bar width=0.155cm, ymin=0, ymax=100,
  ylabel={accuracy on stratum (\%)},
  symbolic x coords={reaction (28),regulatory (11),readiness (7),business (4)},
  xtick=data, tick label style={font=\scriptsize},
  xticklabel style={font=\scriptsize}, label style={font=\footnotesize},
  ymajorgrids, grid style={aubline,very thin}, axis line style={aubgrey},
  legend style={font=\tiny, draw=aubline, fill=white, at={(0.5,1.32)},
                anchor=north, legend columns=3}, legend cell align=left,
]
\addplot[fill=auburl!70, draw=auburl] coordinates {(reaction (28),64.3) (regulatory (11),45.5) (readiness (7),100) (business (4),25)};
\addlegendentry{Qwen3-32B + LoRA-SFT}
\addplot[fill=aubgreen!70, draw=aubgreen] coordinates {(reaction (28),64.3) (regulatory (11),45.5) (readiness (7),100) (business (4),25)};
\addlegendentry{Qwen3-32B + SFT\ensuremath{\to}DPO}
\addplot[fill=auborange!70, draw=auborange] coordinates {(reaction (28),67.9) (regulatory (11),45.5) (readiness (7),85.7) (business (4),0)};
\addlegendentry{Llama-3-8B + LoRA-SFT}
\addplot[fill=aubnavy!70, draw=aubnavy] coordinates {(reaction (28),46.4) (regulatory (11),63.6) (readiness (7),85.7) (business (4),50)};
\addlegendentry{Qwen3-32B}
\addplot[fill=aubgrey!70, draw=aubgrey] coordinates {(reaction (28),46.4) (regulatory (11),63.6) (readiness (7),85.7) (business (4),25)};
\addlegendentry{Qwen3-8B}
\addplot[fill=aubteal!70, draw=aubteal] coordinates {(reaction (28),57.1) (regulatory (11),9.1) (readiness (7),100) (business (4),25)};
\addlegendentry{Qwen3-8B + LoRA-SFT}
\end{axis}\end{tikzpicture}
\caption{\textbf{Accuracy by what actually drove the outcome.} Each Gold-50 case is annotated during verification with the mechanism that decided it: public \emph{reaction}, \emph{regulatory} action, engineering \emph{readiness}, or \emph{business} performance (stratum sizes in parentheses). The ordering is stable across systems: cases decided by regulators or by shipping-readiness are predicted far better than cases decided by how loudly users complained. This is the honest limit of a reaction-simulating pipeline --- the stratum it is architecturally aimed at is the stratum it does worst on. The small strata ($n=7$ and $n=4$) carry very wide intervals and we draw no ranking conclusions from them.}\label{fig:drivers}\end{figure}
}

\newcommand{\figFormatOOD}{%

\begin{figure}[tbp]\centering
\begin{tikzpicture}
\begin{axis}[width=0.99\columnwidth, height=4.6cm,
  ybar, bar width=0.36cm, ymin=0, ymax=92,
  ylabel={accuracy on the same 11 cases (\%)},
  symbolic x coords={{Qwen3-32B base},{Qwen3-8B+SFT},{Qwen3-32B+SFT}},
  xtick=data, tick label style={font=\scriptsize},
  label style={font=\footnotesize}, ymajorgrids,
  grid style={aubline,very thin}, axis line style={aubgrey},
  legend style={font=\scriptsize, draw=aubline, fill=white, at={(0.02,0.98)},
                anchor=north west}, legend cell align=left,
  nodes near coords={\pgfmathprintnumber[fixed,precision=0]{\pgfplotspointmeta}},
  every node near coord/.append style={font=\tiny, anchor=south},
]
\addplot[fill=auborange!45, draw=auborange, point meta=explicit] coordinates {({Qwen3-32B base},18) [18] ({Qwen3-8B+SFT},9) [9] ({Qwen3-32B+SFT},0) [0]};
\addlegendentry{abstract instruction prompt}
\addplot[fill=auburl!70, draw=auburl, point meta=explicit] coordinates {({Qwen3-32B base},54.5) [54.5] ({Qwen3-8B+SFT},63.6) [63.6] ({Qwen3-32B+SFT},72.7) [72.7]};
\addlegendentry{deployed report-stage prompt}
\end{axis}\end{tikzpicture}
\caption{\textbf{The evaluation harness can dominate the measurement.} The same three checkpoints, the same 11 answer-keyed cases, the same scoring code --- only the prompt envelope differs. Asked for a verdict through a generic instruction prompt, the models emit actions outside the five-way schema (\code{delay\_release}, \code{proceed\_with\_mitigations}, \code{rollback}) and accuracy collapses; asked through the exact report-stage prompt the app deploys, the same weights score 54.5--72.7\%. The fine-tuned Qwen3-32B LoRA-SFT (\code{sft\_v3}) is the \emph{most} harness-sensitive of the three, because SFT bound its output format tightly to the serving prompt: it is worst under the abstract prompt and best under the deployed one. We therefore report all headline numbers under the deployed prompt and treat the abstract harness as a schema-compliance probe, not a judgment measurement.}\label{fig:formatood}\end{figure}
}

\newcommand{\figTeacherVsGold}{%

\begin{figure}[tbp]\centering
\begin{tikzpicture}
\begin{axis}[width=0.99\columnwidth, height=5.0cm,
  xlabel={agreement with the cloud teacher's verdict, $n{=}10$ (\%)},
  ylabel={accuracy vs.\ outcome, $n{=}50$ (\%)},
  xmin=0, xmax=100, ymin=0, ymax=80,
  tick label style={font=\scriptsize}, label style={font=\footnotesize},
  grid=both, grid style={aubline,very thin}, axis line style={aubgrey},
  clip mode=individual,
]
\addplot[aubgrey, dashed, thick, mark=none, forget plot] coordinates {(0,0) (80,80)};
\node[font=\tiny,aubgrey,rotate=45,anchor=south] at (axis cs:62,62) {$y=x$};
\addplot[only marks, mark=*, auburl, mark size=1.9] coordinates {(10,56)};
\node[font=\tiny,aubnavy,anchor=west,xshift=5pt,yshift=0pt] at (axis cs:10,56) {Qwen3-32B base};
\addplot[only marks, mark=*, auburl, mark size=1.9] coordinates {(33.3,62)};
\node[font=\tiny,aubnavy,anchor=west,xshift=5pt,yshift=0pt] at (axis cs:33.3,62) {Qwen3-32B+DPO};
\addplot[only marks, mark=*, auburl, mark size=1.9] coordinates {(10,60)};
\node[font=\tiny,aubnavy,anchor=north west,xshift=5pt,yshift=-3pt] at (axis cs:10,60) {Llama-3-8B+SFT};
\addplot[only marks, mark=*, auburl, mark size=1.9] coordinates {(50,50)};
\node[font=\tiny,aubnavy,anchor=north,xshift=0pt,yshift=-5pt] at (axis cs:50,50) {Qwen3-8B+SFT};
\addplot[only marks, mark=*, auburl, mark size=1.9] coordinates {(70,62)};
\node[font=\tiny,aubnavy,anchor=south,xshift=0pt,yshift=5pt] at (axis cs:70,62) {Qwen3-32B+SFT v3};
\end{axis}\end{tikzpicture}
\caption{\textbf{Agreeing with the teacher is not the same as being right.} Horizontal axis: how often a system reproduces the cloud teacher's verdict on the 10 archived runs it was distilled from. Vertical axis: how often the same system matches the verified real-world outcome on the Gold-50 set. Systems sit well below $y=x$, and the teacher itself agrees with itself 10/10 while its own verdicts are not the outcome-correct answer on the cases where both are known (\cref{sec:teacher}). Distillation-style metrics therefore measure imitation, and we use them only as a training-progress signal, never as an accuracy claim.}\label{fig:teachervsgold}\end{figure}
}

\newcommand{\tabFidelity}{%

\begin{table}[t]
\centering\small
\begin{tabular}{llcc}
\toprule
judge & family & direction agree & concern recall \\
\midrule
Opus~4.8        & cloud & 12/16 (75\%) & 67\% \\
GPT-5.2         & cloud & 11/16 (69\%) & 80\% \\
Qwen3-32B       & open  & 12/16 (75\%) & 86\% \\
Gemma-3-27B-it  & open  & 12/16 (75\%) & 90\% \\
\bottomrule
\end{tabular}
\caption{\textbf{Blind fidelity of the synthetic reaction against the real one}, on the
$16$ reaction-driven cases, by four judges across families. Direction agreement is a
floor (see text); \emph{concern recall}---which has no trivial baseline---is the
load-bearing signal and is \emph{higher} for the two open-weight judges.}
\label{tab:fidelity}
\end{table}
}

\newcommand{\tabSimAblation}{%

\begin{table}[t]
\centering\small
\setlength{\tabcolsep}{4pt}
\begin{tabular}{lcccc}
\toprule
solver & \code{nosim} & \code{withsim} & $\Delta$ & $b/c$ \\
\midrule
Opus~4.8        & 84\% & 80\% & $-4\pp$ & 4/6 \\
GPT-5.2         & 46\% & 66\% & $+20\pp$ & 15/5 \\
Qwen3-32B       & 52\% & 56\% & $+4\pp$ & 6/4 \\
Gemma-3-27B-it  & 48\% & 50\% & $+2\pp$ & 7/6 \\
\midrule
\textbf{mean}   &      &      & $\mathbf{+5.5\pp}$ & 3/4 with $b>c$ \\
\bottomrule
\end{tabular}
\caption{\textbf{Matched with/without-simulation ablation}, $50$ cases, exact-match
against the verified outcome. $b$ = cases the simulation fixed (\code{withsim} right,
\code{nosim} wrong); $c$ = cases it broke. Read as within-session deltas: the
\code{withsim} arm reproduces the \Cref{sec:debias} full-context number exactly for GPT
($66\%$), while the gateway Opus run drifted $+12\pp$ across sessions (see text).}
\label{tab:simablation}
\end{table}
}

\newcommand{\figCost}{%

\begin{figure*}[tp]\centering
\begin{tikzpicture}
\begin{axis}[width=0.97\columnwidth, height=6.0cm,
  xlabel={peak VRAM at inference (GB, bf16 $+$ adapter)},
  ylabel={Gold-50 accuracy (\%)},
  xmin=-3, xmax=88, ymin=0, ymax=76,
  ytick={0,20,40,60}, xtick={0,20,40,60,80},
  tick label style={font=\scriptsize}, label style={font=\footnotesize},
  grid=both, grid style={aubline,very thin}, axis line style={aubgrey},
  clip mode=individual,
]
\draw[aubgrey,dashed] (axis cs:-3,20) -- (axis cs:88,20);
\node[font=\scriptsize,aubgrey,anchor=south east] at (axis cs:88,20) {five-way chance, 20\%};
\draw[aubgrey!60,very thin] (axis cs:4.32,46) -- (axis cs:4.96,6);
\draw[aubgrey!60,very thin] (axis cs:17.81,16) -- (axis cs:18.03,60);
\draw[aubgrey!60,very thin] (axis cs:17.7,54) -- (axis cs:18.34,50);
\draw[aubgrey!60,very thin] (axis cs:58.53,36) -- (axis cs:58.66,50);
\draw[aubgrey!60,very thin] (axis cs:67.97,56) -- (axis cs:69.13,62);
\addplot[only marks, mark=square*, auborange, mark size=2.1] coordinates {(4.32,46)};
\node[font=\scriptsize,aubnavy,anchor=west,xshift=4pt,yshift=-3pt] at (axis cs:4.32,46) {Qwen3-1.7B};
\addplot[only marks, mark=*, aubgreen, mark size=2.1] coordinates {(4.96,6)};
\node[font=\scriptsize,aubnavy,anchor=west,xshift=4pt,yshift=0pt] at (axis cs:4.96,6) {Qwen3-1.7B$+$SFT};
\addplot[only marks, mark=square*, auborange, mark size=2.1] coordinates {(17.7,54)};
\node[font=\scriptsize,aubnavy,anchor=west,xshift=4pt,yshift=4pt] at (axis cs:17.7,54) {Qwen3-8B};
\addplot[only marks, mark=square*, auborange, mark size=2.1] coordinates {(17.81,16)};
\node[font=\scriptsize,aubnavy,anchor=west,xshift=4pt,yshift=0pt] at (axis cs:17.81,16) {Llama-3-8B-Instruct};
\addplot[only marks, mark=*, aubgreen, mark size=2.1] coordinates {(18.03,60)};
\node[font=\scriptsize,aubnavy,anchor=west,xshift=4pt,yshift=6pt] at (axis cs:18.03,60) {Llama-3-8B$+$SFT};
\addplot[only marks, mark=*, aubgreen, mark size=2.1] coordinates {(18.34,50)};
\node[font=\scriptsize,aubnavy,anchor=west,xshift=4pt,yshift=-1pt] at (axis cs:18.34,50) {Qwen3-8B$+$SFT};
\addplot[only marks, mark=square*, auborange, mark size=2.1] coordinates {(58.53,36)};
\node[font=\scriptsize,aubnavy,anchor=west,xshift=4pt,yshift=0pt] at (axis cs:58.53,36) {Gemma-3-27B-it};
\addplot[only marks, mark=*, aubgreen, mark size=2.1] coordinates {(58.66,50)};
\node[font=\scriptsize,aubnavy,anchor=west,xshift=4pt,yshift=0pt] at (axis cs:58.66,50) {Gemma-3-27B$+$SFT};
\addplot[only marks, mark=square*, auborange, mark size=2.1] coordinates {(67.97,56)};
\node[font=\scriptsize,aubnavy,anchor=east,xshift=-4pt,yshift=2pt] at (axis cs:67.97,56) {Qwen3-32B};
\addplot[only marks, mark=*, aubgreen, mark size=2.1] coordinates {(68.04,62)};
\node[font=\scriptsize,aubnavy,anchor=south,xshift=0pt,yshift=5pt] at (axis cs:68.04,62) {Qwen3-32B$+$SFT\ensuremath{\to}DPO};
\addplot[only marks, mark=*, aubgreen, mark size=2.1] coordinates {(69.13,62)};
\node[font=\scriptsize,aubnavy,anchor=west,xshift=4pt,yshift=0pt] at (axis cs:69.13,62) {Qwen3-32B$+$SFT};
\end{axis}\end{tikzpicture}
\caption{\textbf{Cost--quality frontier over all eleven systems.} \textcolor{aubgreen}{$\bullet$}~fine-tuned, \textcolor{auborange}{$\blacksquare$}~untuned; grey lines join each backbone to its own adapter. Per-system throughput and latency are in \cref{tab:serving}. All three cost measures are instrumented during the archived-run benchmark; accuracy is the Gold-50 figure. Memory spans $16\times$, from 4.32\,GB to 69.13\,GB, and every system fits on one B200 (183\,GB), so the deployment question is throughput and latency rather than capacity. The frontier is flatter than the leaderboard suggests: the untuned Qwen3-8B base reaches 54\% at 17.7\,GB and 46.6\,tok/s, within 8\,pp of the leader (the Qwen3-32B LoRA-SFT; not significant, $p=0.424$ paired, \cref{sec:leaderboard}) for a quarter of the memory and over three times the throughput. Note that every adapter sits \emph{right-and-slower} than its own base: fine-tuning costs throughput at every scale (\cref{tab:serving}).}\label{fig:cost}\end{figure*}
}

\newcommand{\tabServing}{%

\begin{table}[H]\centering
\caption{Measured inference cost for every system on the leaderboard, from
the archived-run benchmark.
``tok/s'' and ``s/case'' are sequential single-case generation; ``think'' is the
mean count of reasoning tokens emitted before the JSON, which is why the tuned
rows are slower per case than their throughput alone implies. ``wall'' is a
separate measurement: the end-to-end clock for the batched Gold-50 run at
batch 16, so it is not comparable to $\text{s/case}\times50$. Every
configuration fits on one B200.}
\label{tab:serving}
\footnotesize
\setlength{\tabcolsep}{2pt}
\renewcommand{\arraystretch}{1.12}
\begin{tabular}{@{}lcccccc@{}}
\toprule
\textbf{System} & \textbf{Acc} & \textbf{tok/s} & \textbf{VRAM} &
\textbf{s/} & \textbf{think} & \textbf{wall} \\
 & \textbf{\%} & & \textbf{GB} & \textbf{case} & \textbf{tok} &
\textbf{s} \\
\midrule
Qwen3-32B + SFT\ensuremath{\to}DPO & 62 & 17.6 & 68.04 & 133 & 541 & 962 \\
Qwen3-32B + LoRA-SFT & 62 & 13.7 & 69.13 & 178.9 & 604 & 1027 \\
Llama-3-8B + LoRA-SFT & 60 & 28.8 & 18.03 & 278.5 & 544 & 2607 \\
Qwen3-32B & 56 & 24.3 & 67.97 & 90.1 & 517 & 1255 \\
Qwen3-8B & 54 & 46.6 & 17.7 & 40.3 & 527 & 328 \\
Gemma-3-27B + LoRA-SFT & 50 & 12.2 & 58.66 & 177 & 546 & 1618 \\
Qwen3-8B + LoRA-SFT & 50 & 24.1 & 18.34 & 96.8 & 481 & 1963 \\
Qwen3-1.7B & 46 & 59.7 & 4.32 & 27.7 & 412 & 209 \\
Gemma-3-27B-it & 36 & 17.2 & 58.53 & 101.5 & 0 & 535 \\
Llama-3-8B-Instruct & 16 & 62.5 & 17.81 & 127.9 & 0 & 2327 \\
Qwen3-1.7B + LoRA-SFT & 6 & 38 & 4.96 & 182.2 & 647 & 1587 \\
\bottomrule
\end{tabular}
\end{table}
}

\newcommand{\figFtCost}{%

\begin{figure}[tp]\centering
\begin{tikzpicture}
\begin{groupplot}[
  group style={group size=1 by 2, vertical sep=1.5cm,
               xticklabels at=edge bottom},
  width=0.9\columnwidth, height=4.2cm, ybar, /pgf/bar width=0.22cm,
  symbolic x coords={{Qwen3-1.7B},{Llama-3-8B-Instruct},{Qwen3-8B},{Gemma-3-27B-it},{Qwen3-32B}},
  xtick={{Qwen3-1.7B},{Llama-3-8B-Instruct},{Qwen3-8B},{Gemma-3-27B-it},{Qwen3-32B}},
  enlarge x limits=0.16, ymin=0,
  tick label style={font=\tiny},
  xticklabel style={font=\tiny, rotate=32, anchor=north east},
  label style={font=\scriptsize}, title style={font=\scriptsize\bfseries},
  ymajorgrids, grid style={aubline,very thin}, axis line style={aubgrey},
  nodes near coords={\pgfmathprintnumber[fixed,precision=0]{\pgfplotspointmeta}},
  every node near coord/.append style={font=\tiny, anchor=south},
  point meta=explicit,
]
\nextgroupplot[title={(a) Throughput}, ylabel={tokens/s}, ymax=82]
\addplot[fill=auborange!60, draw=auborange] coordinates {({Qwen3-1.7B},59.7) [59.7] ({Llama-3-8B-Instruct},62.5) [62.5] ({Qwen3-8B},46.6) [46.6] ({Gemma-3-27B-it},17.2) [17.2] ({Qwen3-32B},24.3) [24.3]};
\addplot[fill=aubgreen!60, draw=aubgreen] coordinates {({Qwen3-1.7B},38) [38] ({Llama-3-8B-Instruct},28.8) [28.8] ({Qwen3-8B},24.1) [24.1] ({Gemma-3-27B-it},12.2) [12.2] ({Qwen3-32B},13.7) [13.7]};
\nextgroupplot[title={(b) Latency}, ylabel={seconds/case}, ymax=355]
\addplot[fill=auborange!60, draw=auborange] coordinates {({Qwen3-1.7B},27.7) [27.7] ({Llama-3-8B-Instruct},127.9) [127.9] ({Qwen3-8B},40.3) [40.3] ({Gemma-3-27B-it},101.5) [101.5] ({Qwen3-32B},90.1) [90.1]};
\addplot[fill=aubgreen!60, draw=aubgreen] coordinates {({Qwen3-1.7B},182.2) [182.2] ({Llama-3-8B-Instruct},278.5) [278.5] ({Qwen3-8B},96.8) [96.8] ({Gemma-3-27B-it},177) [177] ({Qwen3-32B},178.9) [178.9]};
\end{groupplot}
\end{tikzpicture}
\caption{\textbf{Fine-tuning costs throughput and latency on every backbone we trained.} \textcolor{auborange}{$\blacksquare$}~untuned base, \textcolor{aubgreen}{$\blacksquare$}~the same backbone with its LoRA-SFT adapter. Same recipe, same data, measured on the same hardware. (a) Tokens per second falls for all five backbones --- worst on Llama-3-8B-Instruct, to 46\% of its untuned rate --- because the unmerged LoRA adapter adds work at every forward pass. (b) Seconds per case rises further still, because SFT also taught the models to emit a reasoning trace before the JSON: the two backbones that emitted \emph{no} thinking tokens untuned (Llama-3-8B-Instruct, Gemma-3-27B-it) emit 544 and 546 after it (\cref{tab:serving}). The $+44$\,pp accuracy gain on Llama-3-8B (\cref{tab:paired}) therefore costs 2.2$\times$ the latency per case, which is the trade a deployment actually faces.}\label{fig:ftcost}\end{figure}
}

\newcommand{\figPipeBias}{%

\begin{figure}[tbp]\centering
\begin{tikzpicture}
\begin{groupplot}[
  group style={group size=1 by 2, vertical sep=0.85cm,
               xticklabels at=edge bottom},
  width=0.99\columnwidth, height=3.6cm, ybar, /pgf/bar width=0.42cm,
  symbolic x coords={{Opus 4.8},{GPT-5.2},{Per-agent auto},{Gemma+SFT},{Qwen3-32B},{Qwen3-32B+SFT}},
  xtick={{Opus 4.8},{GPT-5.2},{Per-agent auto},{Gemma+SFT},{Qwen3-32B},{Qwen3-32B+SFT}},
  enlarge x limits=0.13,
  tick label style={font=\scriptsize},
  xticklabel style={font=\tiny, rotate=20, anchor=north east},
  label style={font=\footnotesize},
  title style={font=\small\bfseries, yshift=1pt}, ymajorgrids,
  grid style={aubline,very thin}, axis line style={aubgrey},
  point meta=explicit,
  every axis plot/.append style={bar shift=0pt},
  nodes near coords={\pgfmathprintnumber[fixed,precision=0]{\pgfplotspointmeta}},
  every node near coord/.append style={font=\tiny, anchor=south},
]
\nextgroupplot[title={(a) Accuracy vs verified outcome}, ylabel={\%}, ymin=0, ymax=86, ytick={0,20,40,60,80}]
\addplot[fill=auborange!70, draw=auborange] coordinates {({Opus 4.8},40) [40]};
\addplot[fill=auborange!70, draw=auborange] coordinates {({GPT-5.2},36) [36]};
\addplot[fill=auborange!70, draw=auborange] coordinates {({Per-agent auto},14) [14]};
\addplot[fill=auburl!45, draw=auburl] coordinates {({Gemma+SFT},50) [50]};
\addplot[fill=auburl!45, draw=auburl] coordinates {({Qwen3-32B},56) [56]};
\addplot[fill=auburl!45, draw=auburl] coordinates {({Qwen3-32B+SFT},62) [62]};
\draw[aubgrey,dashed] (rel axis cs:0,0.2326) -- (rel axis cs:1,0.2326);
\node[font=\tiny,aubgrey,anchor=north east] at (rel axis cs:1,0.228) {chance 20\%};
\nextgroupplot[title={(b) Share of cases answered \emph{mitigate}}, ylabel={\% of cases}, ymin=0, ymax=112, ytick={0,20,40,60,80,100}]
\addplot[fill=auborange!70, draw=auborange] coordinates {({Opus 4.8},68) [68]};
\addplot[fill=auborange!70, draw=auborange] coordinates {({GPT-5.2},80) [80]};
\addplot[fill=auborange!70, draw=auborange] coordinates {({Per-agent auto},62) [62]};
\addplot[fill=auburl!45, draw=auburl] coordinates {({Gemma+SFT},64) [64]};
\addplot[fill=auburl!45, draw=auburl] coordinates {({Qwen3-32B},32) [32]};
\addplot[fill=auburl!45, draw=auburl] coordinates {({Qwen3-32B+SFT},28) [28]};
\draw[aubgrey,dashed] (rel axis cs:0,0.1786) -- (rel axis cs:1,0.1786);
\node[font=\tiny,aubgrey,anchor=north east] at (rel axis cs:1,0.174) {true frequency 20\%};
\end{groupplot}\end{tikzpicture}
\caption{\textbf{Adding the simulation stages amplifies over-doom and does not improve the verdict.} \textcolor{auborange}{\textbf{Orange}}: frontier models run through the \emph{complete} five-stage pipeline of \cref{eq:pipeline}. \textcolor{auburl}{\textbf{Teal}}: open-weight models judging the \emph{same} 50 documents at the report stage alone. (a) The full pipeline scores 40\%, 36\%, 14\%; the per-agent routing run falls \emph{below} the $20\%$ chance rate, and all three land beneath every report-only system shown. (b) The mechanism is visible: the full pipeline answers \emph{mitigate} on 68\%, 80\%, 62\% of cases against a true frequency of $20\%$ --- a systematic over-doom bias. The comparison is confounded by model family (frontier vs open-weight), so we read it as evidence about the \emph{bias} and treat the accuracy ordering as the motivation for the matched ablation in \cref{sec:simablation}.}\label{fig:pipebias}\end{figure}
}

\newcommand{\tabCloudFlow}{%

\begin{table}[tbp]\centering
\caption{The Gold-50 set through the \emph{complete} five-stage pipeline under
frontier models, to be read against the report-stage-only results of
\cref{tab:gold50}. Accuracy is reported once a run covers all 50 cases.
\emph{Mitigate} share is given because it is the diagnostic
(\cref{fig:pipebias}); the true frequency is $20\%$. Latency is the median per
case, end to end over all five stages. Regenerated from the run outputs on every
build, so the counts are current as of this document.}
\label{tab:cloudflow}
\small
\setlength{\tabcolsep}{3pt}
\renewcommand{\arraystretch}{1.15}
\begin{tabular}{@{}lcccl@{}}
\toprule
\textbf{Config.} & \textbf{Cases} & \textbf{Median} &
\textbf{\emph{mitig.}} & \textbf{Acc \scriptsize(95\% CI)} \\
 & \textbf{/50} & \textbf{s/case} & & \\
\midrule
Claude Opus 4.8 & 50 & 162 & 68\% & \textbf{40} \scriptsize[27.6,\,53.8] \\
GPT-5.2 & 50 & 150 & 80\% & \textbf{36} \scriptsize[24.1,\,49.9] \\
Per-agent \code{auto} & 50 & 422 & 62\% & \textbf{14} \scriptsize[7.0,\,26.2] \\
\midrule
\textit{best report-only} & 50 & --- & 28\% &
\textit{62.0} \scriptsize[48.2,\,74.1] \\
\bottomrule
\end{tabular}
\end{table}
}

\newcommand{\figDebiasGrid}{%
\begin{figure*}[tp]\centering
\begin{tikzpicture}
\begin{axis}[width=0.9\textwidth, height=9.0cm,
  xbar, bar width=0.20cm, xmin=0, xmax=118, xtick={0,20,40,60,80,100},
  xlabel={accuracy vs.\ verified real-world outcome (\%)},
  y=0.66cm, ymin=-0.8, ymax=14.5,
  ytick={0,1,2,3,4,5,6,7,8,9,10,11,12,13},
  yticklabels={{Qwen3-1.7B + LoRA-SFT},{Llama-3-8B-Instruct},{Qwen3-8B},{Qwen3-8B + LoRA-SFT},{Gemma-3-27B-it},{Qwen3-1.7B},{Gemma-3-27B + LoRA-SFT},{Qwen3-32B},{GPT-5.2},{Claude Opus 4.8},{Llama-3-8B + LoRA-SFT},{Qwen3-32B + SFT\ensuremath{\to}DPO},{Qwen3-32B + LoRA-SFT},{Claude Sonnet 5}},
  yticklabel style={font=\scriptsize}, tick label style={font=\scriptsize},
  label style={font=\footnotesize}, xmajorgrids,
  grid style={aubline,very thin}, axis line style={aubgrey},
  legend style={font=\scriptsize, draw=aubline, fill=white,
                at={(0.98,0.02)}, anchor=south east, row sep=-1pt},
  legend cell align=left, clip mode=individual, enlarge y limits=false,
]
\draw[aubgrey,dashed] (axis cs:20,-0.7) -- (axis cs:20,14);
\node[font=\tiny,aubgrey,anchor=south] at (axis cs:20,14.05) {chance $=20\%$};
\addplot[xbar, fill=aubgrey!30, draw=aubgrey] coordinates {(6,0) (16,1) (54,2) (50,3) (36,4) (46,5) (50,6) (56,7) (38,8) (44,9) (60,10) (62,11) (62,12) (46,13)};
\addlegendentry{baseline prompt}
\addplot[xbar, fill=auburl!75, draw=auburl, error bars/x dir=both, error bars/x explicit] coordinates {(24,0) +- (9.7,13.4) (36,1) +- (11.9,13.9) (40,2) +- (12.4,13.8) (44,3) +- (12.8,13.7) (44,4) +- (12.8,13.7) (48,5) +- (13.2,13.5) (50,6) +- (13.4,13.4) (58,7) +- (13.8,12.6) (66,8) +- (13.8,11.6) (68,9) +- (13.8,11.2) (68,10) +- (13.8,11.2) (76,11) +- (13.4,9.7) (78,12) +- (13.2,9.2) (80,13) +- (13,8.8)};
\addlegendentry{$+$ debias block}
\node[font=\tiny,aubnavy,anchor=west] at (axis cs:37.4,0) {\,24 (+18$^{*}$)};
\node[font=\tiny,aubnavy,anchor=west] at (axis cs:49.9,1) {\,36 (+20$^{*}$)};
\node[font=\tiny,aubnavy,anchor=west] at (axis cs:53.8,2) {\,40 (-14)};
\node[font=\tiny,aubnavy,anchor=west] at (axis cs:57.7,3) {\,44 (-6)};
\node[font=\tiny,aubnavy,anchor=west] at (axis cs:57.7,4) {\,44 (+8)};
\node[font=\tiny,aubnavy,anchor=west] at (axis cs:61.5,5) {\,48 (+2)};
\node[font=\tiny,aubnavy,anchor=west] at (axis cs:63.4,6) {\,50 (+0)};
\node[font=\tiny,aubnavy,anchor=west] at (axis cs:70.6,7) {\,58 (+2)};
\node[font=\tiny,aubnavy,anchor=west] at (axis cs:77.6,8) {\,66 (+28$^{*}$)};
\node[font=\tiny,aubnavy,anchor=west] at (axis cs:79.2,9) {\,68 (+24$^{*}$)};
\node[font=\tiny,aubnavy,anchor=west] at (axis cs:79.2,10) {\,68 (+8)};
\node[font=\tiny,aubnavy,anchor=west] at (axis cs:85.7,11) {\,76 (+14$^{*}$)};
\node[font=\tiny,aubnavy,anchor=west] at (axis cs:87.2,12) {\,78 (+16$^{*}$)};
\node[font=\tiny,aubnavy,anchor=west] at (axis cs:88.8,13) {\,80 (+34$^{*}$)};
\end{axis}\end{tikzpicture}
\caption{\textbf{The debias 2$\times$2: one prompt block, no model change, closes most of the frontier gap.} Every bar is the same open-weight or frontier model on the \emph{same} 50 frozen report-stage contexts through the \emph{same} direct harness (\cref{tab:debias2x2}); grey is the deployed report prompt, \textcolor{auburl}{teal} appends the shared taxonomy-definition $+$ anti-hedge block. Whiskers are 95\% Wilson intervals on the debias arm; the trailing number is that accuracy and, in parentheses, the baseline\ensuremath{\to}debias change, starred when the paired McNemar test is significant at $\alpha{=}0.05$. Defining the taxonomy lifts every frontier model by $+24$ to $+34$\,pp and unwinds the mitigate-collapse (\cref{fig:debiasmech}). The Qwen3-32B LoRA-SFT (\code{sft\_v3}, $78\%$) lands among the frontier: it shows no statistically significant difference from Sonnet ($80\%$, $p{=}1.000$), Opus ($68\%$, $p{=}0.227$) and GPT-5.2 ($66\%$, $p{=}0.180$) once the prompt is fair, having \emph{beaten} all three when it is not (\cref{tab:debiash2h}).}\label{fig:debias2x2}\end{figure*}
}

\newcommand{\tabDebiasGrid}{%

\begin{table*}[tp]\centering
\caption{\textbf{The debias 2$\times$2, every system.} Accuracy vs the verified
outcome on the same 50 frozen contexts and the same direct harness; the only
change is the shared taxonomy-definition $+$ anti-hedge block appended in the
debias arm. $\Delta$ is percentage points; $p$ is the two-sided exact McNemar
test on the paired baseline-vs-debias cases. $\varnothing$ counts unparseable
generations (scored wrong). Generated from the computed metrics.}
\label{tab:debias2x2}
\footnotesize
\setlength{\tabcolsep}{10pt}
\renewcommand{\arraystretch}{1.15}
\begin{tabular}{@{}lccrcc@{}}
\toprule
\textbf{System} & \textbf{Baseline} & \textbf{Debias} & \textbf{$\Delta$} &
\textbf{macro-F1} & \textbf{McNemar} \\
 & \textbf{\%\,\scriptsize[CI]} & \textbf{\%\,\scriptsize[CI]} & \textbf{pp} &
\textbf{b\ensuremath{\to}d} & \textbf{$p$} \\
\midrule \multicolumn{6}{@{}l}{\emph{Frontier (cloud API)}} \\
Claude Sonnet 5 & 46\,\scriptsize[33,59.6] & 80\,\scriptsize[67,88.8] & +34 & 0.43\ensuremath{\to}0.79 & 7.6e-5\,$^{*}$ \\
Claude Opus 4.8 & 44\,\scriptsize[31.2,57.7] & 68\,\scriptsize[54.2,79.2] & +24 & 0.40\ensuremath{\to}0.67 & 0.008\,$^{*}$ \\
GPT-5.2 & 38\,\scriptsize[25.9,51.8] & 66\,\scriptsize[52.2,77.6] & +28 & 0.36\ensuremath{\to}0.64 & 0.003\,$^{*}$ \\
\midrule \multicolumn{6}{@{}l}{\emph{Open-weight, fine-tuned (LoRA)}} \\
Qwen3-32B + LoRA-SFT & 62\,\scriptsize[48.2,74.1] & 78\,\scriptsize[64.8,87.2] & +16 & 0.59\ensuremath{\to}0.78 & 0.039\,$^{*}$ \\
Qwen3-32B + SFT\ensuremath{\to}DPO & 62\,\scriptsize[48.2,74.1] & 76\,\scriptsize[62.6,85.7]\,\scriptsize{1$\varnothing$} & +14 & 0.59\ensuremath{\to}0.77 & 0.039\,$^{*}$ \\
Llama-3-8B + LoRA-SFT & 60\,\scriptsize[46.2,72.4]\,\scriptsize{1$\varnothing$} & 68\,\scriptsize[54.2,79.2] & +8 & 0.58\ensuremath{\to}0.62 & 0.344 \\
Gemma-3-27B + LoRA-SFT & 50\,\scriptsize[36.6,63.4] & 50\,\scriptsize[36.6,63.4]\,\scriptsize{2$\varnothing$} & +0 & 0.44\ensuremath{\to}0.46 & 1.000 \\
Qwen3-8B + LoRA-SFT & 50\,\scriptsize[36.6,63.4]\,\scriptsize{6$\varnothing$} & 44\,\scriptsize[31.2,57.7]\,\scriptsize{1$\varnothing$} & -6 & 0.51\ensuremath{\to}0.40 & 0.607 \\
Qwen3-1.7B + LoRA-SFT & 6\,\scriptsize[2.1,16.2]\,\scriptsize{35$\varnothing$} & 24\,\scriptsize[14.3,37.4]\,\scriptsize{17$\varnothing$} & +18 & 0.07\ensuremath{\to}0.23 & 0.035\,$^{*}$ \\
\midrule \multicolumn{6}{@{}l}{\emph{Open-weight, untuned base}} \\
Qwen3-32B & 56\,\scriptsize[42.3,68.8] & 58\,\scriptsize[44.2,70.6] & +2 & 0.56\ensuremath{\to}0.52 & 1.000 \\
Qwen3-1.7B & 46\,\scriptsize[33,59.6] & 48\,\scriptsize[34.8,61.5] & +2 & 0.39\ensuremath{\to}0.38 & 1.000 \\
Gemma-3-27B-it & 36\,\scriptsize[24.1,49.9] & 44\,\scriptsize[31.2,57.7] & +8 & 0.34\ensuremath{\to}0.34 & 0.481 \\
Qwen3-8B & 54\,\scriptsize[40.4,67] & 40\,\scriptsize[27.6,53.8] & -14 & 0.52\ensuremath{\to}0.28 & 0.065 \\
Llama-3-8B-Instruct & 16\,\scriptsize[8.3,28.5] & 36\,\scriptsize[24.1,49.9] & +20 & 0.16\ensuremath{\to}0.30 & 0.006\,$^{*}$ \\
\bottomrule
\end{tabular}\end{table*}
}

\newcommand{\tabDebiasHead}{%

\begin{table*}[tp]\centering
\caption{\textbf{Does the open fine-tune beat the frontier? A paired
question.} Qwen3-32B + LoRA-SFT (\code{sft\_v3}) against each
frontier model on the \emph{same} 50 cases, so the correct test is a two-sided
exact McNemar on the discordant pairs, not a difference of accuracies --- the
Wilson intervals overlap in every pair. This is the paper's \emph{primary}
hypothesis family; $p_{\text{Holm}}$ is the Holm-corrected value across the three
comparisons within each arm. Under the under-specified baseline prompt the
Qwen3-32B LoRA-SFT wins all three, significant after correction; once the
taxonomy is defined no significant difference from any frontier model is detected
(``n.s.''), which is a failure to detect a difference at $n{=}50$, not evidence
of equivalence. Generated from the computed metrics.}
\label{tab:debiash2h}
\footnotesize
\setlength{\tabcolsep}{5pt}
\renewcommand{\arraystretch}{1.1}
\begin{tabular}{@{}llcccc@{}}
\toprule
\textbf{Arm} & \textbf{vs.\ frontier} & \textbf{Acc (ft\,/\,cloud)} &
\textbf{Wins (ft--cloud)} & \textbf{$p$} & \textbf{$p_{\text{Holm}}$} \\
\midrule
\multirow{3}{*}{Baseline} & Claude Sonnet 5 & 62\,/\,46 & 10--2 & 0.039 & 0.039 (\textbf{beats}) \\
 & Claude Opus 4.8 & 62\,/\,44 & 10--1 & 0.012 & 0.023 (\textbf{beats}) \\
 & GPT-5.2 & 62\,/\,38 & 13--1 & 0.002 & 0.005 (\textbf{beats}) \\
\midrule
\multirow{3}{*}{Debias} & Claude Sonnet 5 & 78\,/\,80 & 4--5 & 1.000 & 1.000 (n.s.) \\
 & Claude Opus 4.8 & 78\,/\,68 & 8--3 & 0.227 & 0.539 (n.s.) \\
 & GPT-5.2 & 78\,/\,66 & 10--4 & 0.180 & 0.539 (n.s.) \\
\bottomrule
\end{tabular}\end{table*}
}

\newcommand{\figDebiasMech}{%

\begin{figure}[tbp]\centering
\begin{tikzpicture}
\begin{groupplot}[
  group style={group size=2 by 1, horizontal sep=1.5cm},
  width=0.30\columnwidth, height=4.9cm,
  tick label style={font=\scriptsize}, label style={font=\footnotesize},
  title style={font=\scriptsize\bfseries, yshift=-2pt},
  enlarge x limits=false, enlarge y limits=false, y dir=reverse,
  ytick={0,1,2,3,4},
  yticklabels={{Claude Sonnet 5},{Claude Opus 4.8},{GPT-5.2},{Qwen3-32B + LoRA-SFT},{Qwen3-32B + SFT\ensuremath{\to}DPO}},
  xtick={0,1}, xticklabels={baseline, debias},
  xticklabel style={rotate=25, anchor=north east},
  y=0.66cm, x=1.02cm,
]
\nextgroupplot[title={(a) \emph{mitigate} preds}, colormap={aub2}{color=(white) color=(aubpeach) color=(auborange)}, point meta min=0, point meta max=40]
\addplot[matrix plot*, mesh/cols=2, point meta=explicit] table[meta=C, row sep=crcr] {x y C\\0 0 36\\1 0 16\\0 1 33\\1 1 18\\0 2 35\\1 2 19\\0 3 14\\1 3 16\\0 4 14\\1 4 18\\};
\node[font=\tiny,white] at (axis cs:0,0) {36};
\node[font=\tiny,aubnavy] at (axis cs:1,0) {16};
\node[font=\tiny,white] at (axis cs:0,1) {33};
\node[font=\tiny,aubnavy] at (axis cs:1,1) {18};
\node[font=\tiny,white] at (axis cs:0,2) {35};
\node[font=\tiny,aubnavy] at (axis cs:1,2) {19};
\node[font=\tiny,aubnavy] at (axis cs:0,3) {14};
\node[font=\tiny,aubnavy] at (axis cs:1,3) {16};
\node[font=\tiny,aubnavy] at (axis cs:0,4) {14};
\node[font=\tiny,aubnavy] at (axis cs:1,4) {18};
\nextgroupplot[title={(b) \emph{segment} recall}, yticklabels={}, colormap={aub}{color=(white) color=(aubteal) color=(auburl)}, point meta min=0, point meta max=10]
\addplot[matrix plot*, mesh/cols=2, point meta=explicit] table[meta=C, row sep=crcr] {x y C\\0 0 1\\1 0 10\\0 1 1\\1 1 7\\0 2 1\\1 2 6\\0 3 2\\1 3 8\\0 4 2\\1 4 7\\};
\node[font=\tiny,aubnavy] at (axis cs:0,0) {1};
\node[font=\tiny,white] at (axis cs:1,0) {10};
\node[font=\tiny,aubnavy] at (axis cs:0,1) {1};
\node[font=\tiny,white] at (axis cs:1,1) {7};
\node[font=\tiny,aubnavy] at (axis cs:0,2) {1};
\node[font=\tiny,white] at (axis cs:1,2) {6};
\node[font=\tiny,aubnavy] at (axis cs:0,3) {2};
\node[font=\tiny,white] at (axis cs:1,3) {8};
\node[font=\tiny,aubnavy] at (axis cs:0,4) {2};
\node[font=\tiny,white] at (axis cs:1,4) {7};
\end{groupplot}\end{tikzpicture}
\caption{\textbf{What the block changes.} For the five capable systems, baseline vs debias. (a) The over-prediction of \emph{mitigate} against a true frequency of 10/50 collapses for the frontier models (Sonnet $36{\to}16$, Opus $33{\to}18$, GPT $35{\to}19$); the fine-tunes never had it (\code{sft\_v3} $14{\to}16$), so their gain comes elsewhere. (b) It comes from \emph{segment}: the class that was nearly unreachable ($1$--$2$/10) is recovered to $6$--$10$/10 once the taxonomy names what \emph{segment} means. Fine-tuning and the prompt block therefore fix overlapping but distinct failures --- \cref{tab:debias2x2} gives all fourteen systems.}\label{fig:debiasmech}\end{figure}
}